\documentclass[twoside,11pt]{article}

\usepackage{jmlr2e}
\usepackage{lipsum}
\usepackage{amsmath}
\usepackage{bm}
\usepackage{hyperref}
\usepackage{xcolor}
\usepackage{float}
\usepackage{pifont}% http://ctan.org/pkg/pifont
\usepackage{graphicx}
\usepackage{caption}
\usepackage{subcaption}

\hypersetup{
     colorlinks=true,
     linkcolor=black,
     filecolor=blue,
     citecolor = black,      
     urlcolor=black,
     }

\newcommand{\xmark}{\ding{55}}

\ShortHeadings{Efficient Transformers: A Survey}{Efficient Transformers: A Survey}
\firstpageno{1}

\begin{document}

\title{Efficient Transformers: A Survey}

\author{\name Yi Tay \email yitay@google.com
\addr \\ Google Research
    %   \\Mountain View, California
\AND 
\name Mostafa Dehghani \email dehghani@google.com 
\addr \\ Google Research, Brain team
    %   Amsterdam, Netherlands
\AND 
\name Dara Bahri \email dbahri@google.com 
\addr \\ Google Research
    %   \\ Mountain View, California
\AND
\name Donald Metzler \email metzler@google.com
       \addr \\ Google Research 
    %   \\ Mountain View, California
        }

\editor{Preprint, Version 2, Updated Mar 2022}

% Math helper definitions
\def\reals{\mathbb{R}}

\maketitle

\begin{abstract}%   <- trailing '%' for backward compatibility of .sty file
Transformer model architectures have garnered immense interest lately due to their effectiveness across a range of domains like language, vision and reinforcement learning. In the field of natural language processing for example, Transformers have become an indispensable staple in the modern deep learning stack. Recently, a dizzying number of \emph{``X-former''} models have been proposed - Reformer, Linformer, Performer, Longformer, to name a few - which improve upon the original Transformer architecture, many of which make improvements around computational and memory \emph{efficiency}. With the aim of helping the avid researcher navigate this flurry, this paper characterizes a large and thoughtful selection of recent efficiency-flavored ``X-former'' models, providing an organized and comprehensive overview of existing work and models across multiple domains.
\end{abstract}

% TODOs for the next updated version:

% P0:
% apply comments from big bird folk (Avi, Manzil, Guru)
% apply Tamas comments
% cite funnel transformer --> DONE
% cite adaptive span attention and adaptively spares transformer --> DONE
% cite video transformer --> DONE
% cite transformer with clustered attention --> DONE

% P1:
% add  a section for adaptive span attention / adaptively sparse transformer
% maybe add Fast Transformers with Clustered Attention 

\section{Introduction}
Transformers~\citep{vaswani2017attention} are a formidable force in the modern deep learning stack. Transformers are pervasive and have made tremendous impact in many fields such as language understanding~\citep{devlin2018bert,brown2020language,raffel2019exploring} and image processing~\citep{parmar2018image,carion2020end}. As such, it is only natural that a wealth of research has been dedicated to making fundamental improvements to the model over the past few years~\citep{dehghani2018universal,so2019evolved,ahmed2017weighted}. This immense interest has also spurred research into more efficient variants of the model~\citep{kitaev2020reformer,roy2020efficient,beltagy2020longformer,katharopoulos2020transformers,tay2020sparse,wang2020linformer,rae2020compressive,choromanski2020rethinking,dai2020funnel,correia2019adaptively, sukhbaatar2019adaptive, vyas2020fast}.

There has been such a surge of Transformer model variants proposed recently, that researchers and practitioners alike may find it challenging to keep pace with the rate of innovation. As of this writing and this manuscript's first draft (circa August 2020), there have been nearly a dozen new efficiency-focused models proposed in just the past 6 months. Thus, a survey of the existing literature is both beneficial for the community and quite timely.

The self-attention mechanism is a key defining characteristic of Transformer models. The mechanism can be viewed as a graph-like inductive bias that connects all tokens in a sequence with a relevance-based pooling operation. A well-known concern with self-attention is the quadratic time and memory complexity, which can hinder model scalability in many settings. There has been an overwhelming influx of model variants proposed recently that address this problem. We hereinafter name this class of models \textit{``efficient Transformers''}. 

The \emph{efficiency} of a model can be interpreted in a variety of ways. It might refer to the memory footprint of the model, which is of importance when the memory of accelerators on which the model is running is limited. Efficiency might also refer to computational costs, e.g. the number of FLOPs, both during training and inference. In particular, for on-device applications, models often must operate within a highly constrained computational budget. Throughout this survey, we refer to the efficiency of Transformers both in terms of memory and computation. We are especially interested in how such models perform when they are applied to large inputs.  

Efficient self-attention models are crucial in applications that model long sequences. For example, documents, images, and videos are all often composed of a relatively large number of pixels or tokens. Efficiency in processing long sequences is therefore paramount for widespread adoption of Transformers.

This survey sets out to provide a comprehensive overview of the recent advances made in this class of models. We are primarily interested in modeling advances and architectural innovations that improve the general efficiency of Transformers, including but not limited to tackling the quadratic complexity issue of the self-attention mechanism or reducing the computation costs by means such as pooling and/or sparsity. We also briefly discuss general improvements and other efficiency improvements such as parameter sharing.

We propose a taxonomy of efficient Transformer models, characterizing them by their technical innovation and primary use case. Specifically, we review Transformer models that have applications in both language and vision domains, attempting to consolidate the literature across the spectrum. We also provide a detailed walk-through of many of these models and draw connections between them.

\paragraph{Author notes on the updated version (December 2021)} This manuscript went through a round of revision in December 2021 (approximately a year and 4 months later after the first manuscript was written). The main changes involve adding our discussions to better reflect the state of research at this current point of time (new models, new paradigms) and also accurately reflect the current meta trends surrounding this research area. A retrospective section is posed near the end of the paper. See Appendix for a meaningful change log of what has happened as we transitioned to V2 of this survey.

\paragraph{Author notes on the updated version (March 2022)} We wanted to post the update to arxiv in Jan but forgot about it. We lightly revised it again in Mar by adding newer SOTA sparse models such as ST-MoE-32B \citep{zoph2022designing}.

\begin{figure}[t!]
    \centering
    \includegraphics[width=1.0\linewidth]{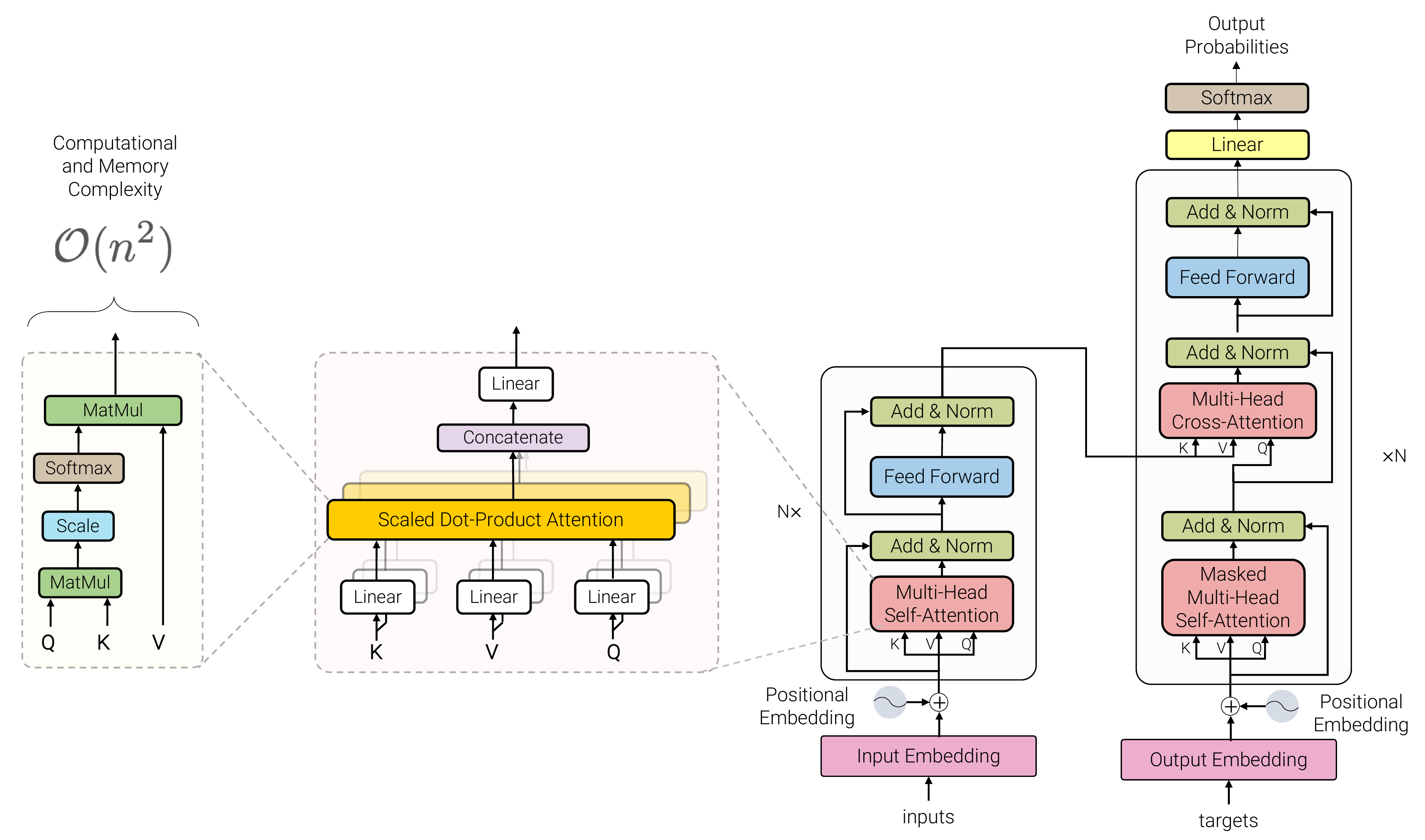}
        \vspace{-2em}
    \label{fig:transformer_arch}
\caption{Architecture of the standard Transformer~\citep{vaswani2017attention}}
\label{stats}
\end{figure}

\section{Background on Transformers}
This section provides an overview of the well-established Transformer architecture~\citep{vaswani2017attention}. Transformers are multi-layered architectures formed by stacking Transformer blocks on top of one another. 

Transformer blocks are characterized by a multi-head self-attention mechanism, a position-wise feed-forward network, layer normalization~\citep{ba2016layer} modules and residual connectors. The input to the Transformer model is often a tensor of shape $\reals^B \times \reals^N$, where $B$ is the batch size, $N$ the sequence length.

The input first passes through an embedding layer that converts each one-hot token representation into a $d_{model}$ dimensional embedding, i.e., $\reals^B \times \reals^N \times \reals^{d_{model}}$. The new tensor is then additively composed with positional encodings and passed through a multi-headed self-attention module. Positional encodings can take the form of a sinusoidal input (as per~\citep{vaswani2017attention}) or be trainable embeddings.

The inputs and output of the multi-headed self-attention module are connected by residual connectors and a layer normalization layer. The output of the multi-headed self-attention module is then passed to a two-layered feed-forward network which has its inputs/outputs similarly connected in a residual fashion with layer normalization. The sub-layer residual connectors with layer norm is expressed as:
\begin{align*}
X = \text{LayerNorm}(F_S(X)) + X    
\end{align*}
where $F_S$ is the sub-layer module which is either the multi-headed self-attention or the position-wise feed-forward layers.

% For the rest of this technical section, we define $N$ as the sequence length, $d$ as the hidden dimension size of the Transformer.

\subsection{Multi-Head Self-Attention} The Transformer model leverages a multi-headed self-attention mechanism. The key idea behind the mechanism is for each element in the sequence to learn to gather from other tokens in the sequence. The operation for a single head is defined as:
\begin{align*}
A_h =  \text{Softmax}(\alpha Q_hK_h^\top)V_h, 
\end{align*}
where $X$ is a matrix in $\reals^{N \times d}$, $\alpha$ is a scaling factor that is typically set to $\frac{1}{\sqrt{d}}$, $Q_h=X\bm{W}_q, K_h=X\bm{W}_k$ and $V_h=X\bm{W}_v$ are linear transformations applied on the temporal dimension of the input sequence, $\bm{W}_q, \bm{W}_k, \bm{W}_v \in \mathbb{R}^{d \times \frac{d}{H}}$ are the weight matrices (parameters) for the query, key, and value projections that project the input $X$ to an output tensor of $d$ dimensions, and $N_H$ is the number of heads. Softmax is applied row-wise.

The outputs of heads $A_1 \cdots A_{H}$ are concatenated together and passed into a dense layer. The output $Y$ can thus be expressed as $Y = \bm{W}_o[A_1 \cdots A_{H}]$,  where $\bm{W}_{o}$ is an output linear projection. Note that the computation of $A$ is typically done in a parallel fashion by considering tensors of $\reals^B \times \reals^N \times \reals^{H}  \times \reals^{\frac{d}{H}}$ and computing the linear transforms for all heads in parallel.

The attention matrix $A=QK^\top$ is chiefly responsible for learning alignment scores between tokens in the sequence. In this formulation, the dot product between each element/token in the query ($Q$) and key ($K$) is taken. This drives the self-alignment process in self-attention whereby tokens learn to \textit{gather} from each other.

\subsection{Position-wise Feed-forward Layers} The outputs of the self-attention module are then passed into a two-layered feed-forward network with ReLU activations. This feed-forward layer operates on each position independently. This is expressed as follows:
\begin{align*}
F_2(ReLU(F_1(X_A)))    
\end{align*}
where $F_1$ and $F_2$ are feed-forward functions of the form $Wx +b$. 

\subsection{Putting it all together}
Each Transformer block can be expressed as:
\begin{align*}
X_A &= \text{LayerNorm}(\text{MultiheadAttention}(X, X)) + X\\
X_B &= \text{LayerNorm}(\text{PositionFFN}(X_A)) + X_A 
\end{align*}
where $X$ is the input of the Transformer block and $X_B$ is the output of the Transformer block. Note that the $\text{MultiheadAttention}()$ function accepts two argument tensors, one for query and the other for key-values. If the first argument and second argument is the same input tensor, this is the \text{MultiheadSelfAttention} mechanism.

\subsection{On the compute cost of Transformers} The computation costs of Transformers is derived from multiple factors. Firstly, the memory and computational complexity required to compute the attention matrix is quadratic in the input sequence length, i.e., $N \times N$. In particular, the $QK^\top$ matrix multiplication operation alone consumes $N^2$ time and memory. This restricts the overall utility of self-attentive models in applications which demand the processing of long sequences. Memory restrictions are tend to be applicable more to training (due to gradient updates) and are generally of lesser impact on inference (no gradient updates). The quadratic cost of self-attention impacts speed$\footnote{We would like to emphasize that complexity does not always translate to real world throughput or latency. A model of linear complexity can be slower than a model with quadratic complexity in practice.}$ in both training and inference. The compute costs of the self-attention mechanism contributes partially to the overall compute cost of the Transformer. A non-trivial amount of compute still stems from the two layer feed-forward layers at every Transformer block (approximately half the compute time and/or FLOPs). The complexity of the FFN is linear with respect to sequence length but is generally still costly. Hence, a large portion of recent work have explored sparsity~\citep{lepikhin2020gshard,fedus2021switch} as a means to scale up the FFN without incurring compute costs. Efficient attention and efficient models are generally orthogonal - although some efficient attention methods explicitly aim to reduce the sequence length~\citep{dai2020funnel} and as a result also save computation costs in both aspects. Efficiency and computational costs is generally a complicated affair and we would suggest readers peruse~\citep{dehghani2021efficiency} for more details on trade-offs, intricacies etc.

% In contrast, the feed-forward computation is only linear in $N$, since the \emph{same}
% feed-forward layer is applied to the output of the self-attention layer exactly once per sequence position. As a result, most efficient Transformer methods target the attention computation and less so the other ones, like the feed-forward computation.
% , in hopes of improving the scalability of the Transformer model.

\subsection{Transformer Mode} It is important to note the differences in how the Transformer blocks are used. Transformers can primarily be used in three ways, namely: (1) \emph{encoder-only} (e.g., for classification), (2) \emph{decoder-only} (e.g., for language modeling), and (3) \emph{encoder-decoder} (e.g., for machine translation). In encoder-decoder mode, there are usually multiple multi-headed self-attention modules, including a standard self-attention in both the encoder and the decoder, along with an encoder-decoder cross-attention that allows the decoder to utilize information from the encoder.
This influences the design of the self-attention mechanism. In the encoder mode, there is no restriction or constraint that the self-attention mechanism has to be causal, i.e., dependent solely on the present and past tokens. In the encoder-decoder setting, self-attention used in the decoder (i.e. across decoding positions) must be causal since each auto-regressive decoding step can only depend on previous tokens, whereas the self-attention used in the encoder need not. Fulfilling this requirement can prove challenging for many efficient self-attention designs.

The mode of usage of a Transformer model generally depends on the target application. Given an input sequence, the sequence is typically passed through an encoder stack. At this stage, there might be too options. For multi-class classification, a linear layer with Softmax outputs typically projects the sequence representation down to the number of classes. In the case of BERT~\citep{devlin2018bert}, this is a \textit{[CLS]} token that is appended to the start of the sequence as a prefix. Recent work has also explored the usage of Encoder-Decoder architectures for classification, such as T5~\citep{raffel2019exploring}. Decoder-only models are typically used for generation and are trained using a language modeling objective (of predicting the next token). Due to the nature of the loss, these models are often superior for open ended generation~\citep{brown2020language}. A decoder-only model needs to be causal and a upper triangular mask needs to be applied to prevent tokens from peeping into the future. We refer interested readers to~\citep{raffel2019exploring} for more detailed descriptions of the various Transformer modes.

\subsection{Applications}
Transformers have a wide range of applications ranging from language to vision, speech and reinforcement learning. It was initially introduced within the context of sequence to sequence machine translation in NLP. Following which, most of the applications of Transformers have been within the context of language - given the concurrent advance of pretrained models such as BERT~\citep{devlin2018bert}. Many early improvements to this line of efficient transformers is therefore focused on language processing applications~\citep{beltagy2020longformer,ainslie2020etc}. For historical reasons, this survey paper leans slightly towards language. However, it is also worth noting that a substantial amount of papers considered in our survey also considers multimodal applications whereby a sequence processor is required. For example \citet{roy2020efficient,choromanski2020rethinking,tay2020sparse,child2019generating} considers generative modeling task on images or other modalities such as proteins. 
\begin{figure}[t]
    \centering
    \includegraphics[width=1.0\linewidth]{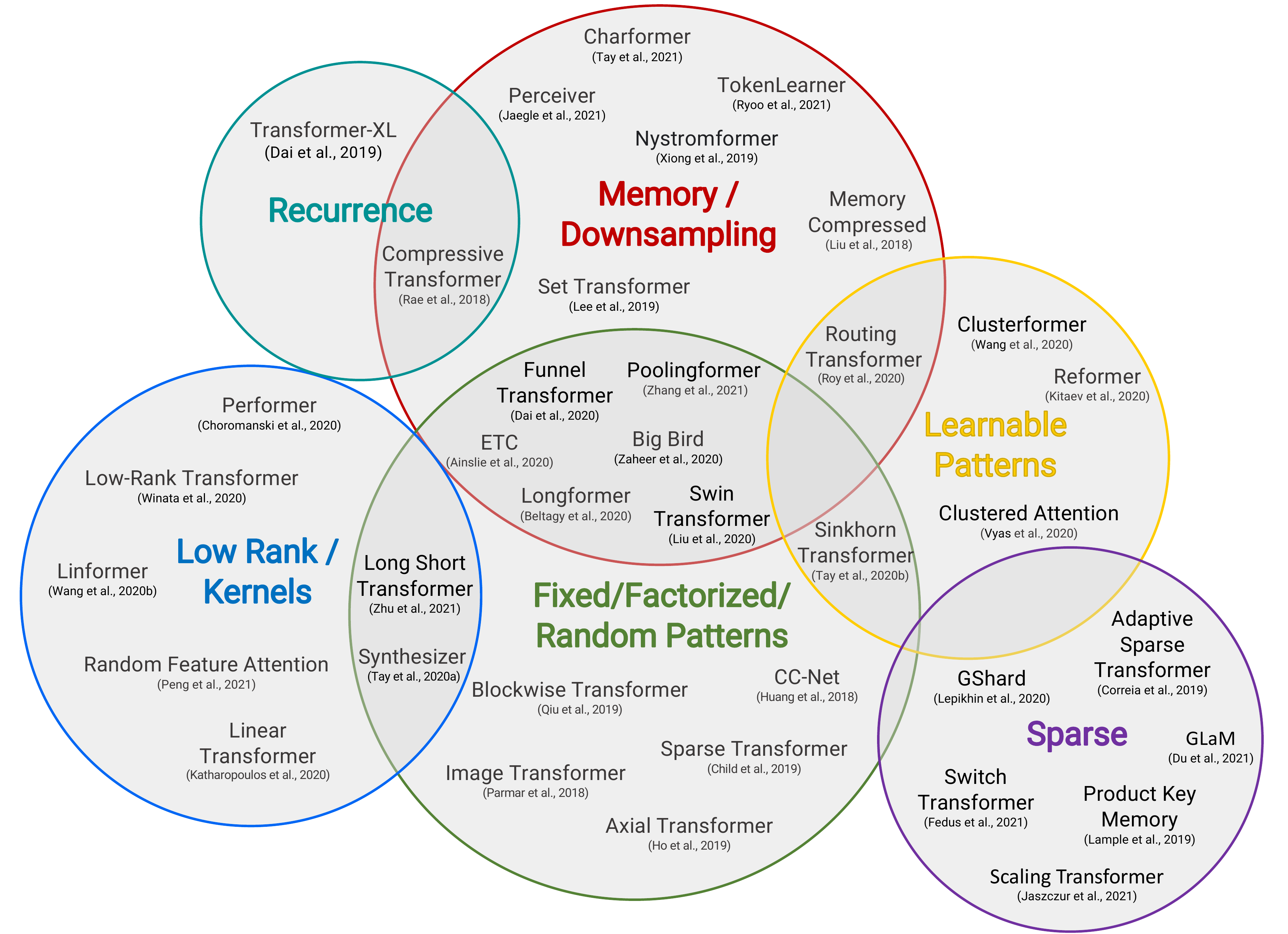}
    \caption{Taxonomy of Efficient Transformer Architectures.}
    \label{fig:taxonomy}
\end{figure}
\section{A Survey of Efficient Transformer Models}

In this section, we provide a high-level overview of efficient Transformer models. We begin by presenting a characterization of the different models. Table~\ref{tab:summarytable} lists the efficient Transformers released to date while Figure \ref{fig:taxonomy} presents a graphical overview of several key efficient Transformer models.

\begin{table}[t]
    \centering
    \small
    \begin{tabular}{l|c|c|l}
    \hline
       Model / Paper  &  Complexity & Decode & Class\\
       \hline
       Memory Compressed~\citep{liu2018generating} & $\mathcal{O}(N_c^2)$& $\checkmark$ &  FP+M \\
        Image Transformer~\citep{parmar2018image} & $\mathcal{O}(N.m)$& $\checkmark$ & FP \\
        % Music Transformer$^\dagger$~\citep{huang2018music} & & & \\
        Set Transformer~\citep{lee2019set} & $\mathcal{O}(kN)$& $\text{\xmark}$  & M\\ 
         Transformer-XL~\citep{dai2019transformer} & $\mathcal{O}(N^2)$ & $\checkmark$ & RC \\ 
        %   Star Transformer$^\dagger$~\citep{guo2019star} & $\mathcal{O}(nm)$ & $\text{\xmark}$ & M \\
        Sparse Transformer   ~\citep{child2019generating} & $\mathcal{O}(N \sqrt{N})$ &  $\checkmark$& FP\\
        % Adaptive Attn. Span$^\dagger$~\citep{sukhbaatar2019adaptive} & & & \\
      
        % + Factorized \\
        %  Blockwise Transformer~\citep{qiu2019blockwise}& $\mathcal{O}(b^2)$ & $\checkmark$ & FP\\
        Reformer~\citep{kitaev2020reformer} & $\mathcal{O}(N \log N)$ & $\checkmark$ & LP\\
        Routing Transformer~\citep{roy2020efficient} & $\mathcal{O}(N\sqrt{N)}$ & $\checkmark$ & LP \\
         Axial Transformer~\citep{ho2019axial} & $\mathcal{O}(N \sqrt{N})$ & $\checkmark$ & FP \\
         Compressive Transformer~\citep{rae2020compressive} & $\mathcal{O}(N^2)$ & $\checkmark$& RC \\
        %  BP-Transformer~\citep{ye2019bp} & & & \\
        Sinkhorn Transformer~\citep{tay2020sparse} & $\mathcal{O}(B^2)$ & $\checkmark$  & LP \\
        % Sparse Adaptive Connection~\citep{li2020sac} & $\mathcal{O}(kN)$ & & \\
        Longformer~\citep{beltagy2020longformer} & $\mathcal{O}(n(k+m))$ & $\checkmark$ & FP+M\\ 
        ETC~\citep{ainslie2020etc} & $\mathcal{O}(N_g^2 + N N_g)$ &  $\text{\xmark}$ & FP+M\\ 
        
        Synthesizer~\citep{tay2020synthesizer} & $\mathcal{O}(N^2)$ & $\checkmark$ & LR+LP\\
          Performer~\citep{choromanski2020masked} & $\mathcal{O}(N)$ & $\checkmark$ & KR\\
        %   GMAT~\citep{gupta2020gmat} & $\mathcal{O}(k(n+k))$& $\text{\xmark}$ & FP + M \\
        Funnel Transformer~\citep{dai2020funnel} & $\mathcal{O}(N^2)$ & $\checkmark$ & FP+DS\\
           Linformer~\citep{wang2020linformer}  & $\mathcal{O}(N)$ & $\text{\xmark}$ & LR \\
        Linear Transformers ~\citep{katharopoulos2020transformers} & $\mathcal{O}(N)$ & $\checkmark$ & KR \\
    % Fast Transformers~\citep{vyas2020fast} & $\mathcal{O}(kN)$ & & \\ 
        Big Bird~\citep{zaheer2020big} &$\mathcal{O}(N)$ &$\text{\xmark}$ & FP+M\\
        Random Feature Attention~\citep{peng2021random} & $\mathcal{O}(N)$ & $\checkmark$ & KR\\
        Long Short Transformers~\citep{zhu2021long} &$\mathcal{O}(kN)$ & $\checkmark$ & FP + LR \\ 
        Poolingformer~\citep{zhang2021poolingformer} & $\mathcal{O}(N)$ & $\text{\xmark}$   & FP+M \\
        Nystr\"{o}mformer~\citep{xiong2021nystr} & $\mathcal{O}(kN)$ &  $\text{\xmark}$ & M+DS\\
          Perceiver~\citep{jaegle2021perceiver} &$\mathcal{O}(kN)$ &$\checkmark$ & M+DS \\ 
        Clusterformer~\citep{wang2020cluster} & $\mathcal{O}(N\log N)$ & $\text{\xmark}$ & LP \\
        Luna~\citep{ma2021luna} &  $\mathcal{O}(kN)$ &$\text{\checkmark}$ & M\\
        TokenLearner~\citep{ryoo2021tokenlearner} & $\mathcal{O}(k^2)$ & $\text{\xmark}$ &DS \\
         \hline
        Adaptive Sparse Transformer~\citep{correia2019adaptively} & $\mathcal{O}(N^2)$ & $\text{\checkmark}$ & Sparse \\
        Product Key Memory~\citep{lample2019large} & $\mathcal{O}(N^2)$ & $\checkmark$& Sparse\\
        Switch Transformer~\citep{fedus2021switch} & $\mathcal{O}(N^2)$& $\checkmark$ & Sparse\\
        ST-MoE~\citep{zoph2022designing} & $\mathcal{O}(N^2)$& $\checkmark$ & Sparse\\
        GShard~\citep{lepikhin2020gshard} & $\mathcal{O}(N^2)$ & $\checkmark$ & Sparse \\
        Scaling Transformers~\citep{jaszczur2021sparse} & $\mathcal{O}(N^2)$ &$\checkmark$ & Sparse\\
        GLaM~\citep{du2021glam}  & $\mathcal{O}(N^2)$ &$\checkmark$ & Sparse\\ 
        \hline
    \end{tabular}
    \caption{Summary of Efficient Transformer Models. Models in the first section are mainly efficient attention methods. Models in the subsequent lower section generally refer to sparse models. Class abbreviations include: FP = Fixed Patterns or Combinations of Fixed Patterns, M = Memory, LP = Learnable Pattern, LR = Low-Rank, KR = Kernel RC = Recurrence, and DS = Downsampling. Furthermore, $N$ generally refers to the sequence length and $B$ is the local window (or block) size. $N_g$ and $N_c$ denote global model memory length and convolutionally-compressed sequence lengths respectively.}
    \label{tab:summarytable}
\end{table}

\subsection{A Taxonomy of Efficient Transformers}
This section outlines a general taxonomy of efficient Transformer models, characterized by their core techniques and primary use case. While, the primary goal of most of these models is to improve the memory complexity if the self-attention mechanism, we also include methods that improve the general efficiency of the Transformer architecture.
% in contrast to the vanilla Transformer that incorporates a self-attention mechanism which is dense in nature.

\begin{itemize}
\item \textbf{Fixed Patterns (FP)} - The earliest modifications to self-attention simply sparsifies the attention matrix by limiting the field of view to fixed, predefined patterns such as local windows and block patterns of fixed strides.
\begin{itemize}
    \item \textbf{Blockwise Patterns} The simplest example of this technique in practice is the blockwise (or chunking) paradigm which considers blocks of local receptive fields by chunking input sequences into fixed blocks. Examples of models that do this include Blockwise~\citep{qiu2019blockwise} and/or Local Attention~\citep{parmar2018image}. Chunking input sequences into blocks reduces the complexity from $N^2$ to $B^2$ (block size) with $B<<N$, significantly reducing the cost. These blockwise or chunking methods serve as a basis for many more complex models. 
    \item \textbf{Strided Patterns} Another approach is to consider strided attention patterns, i.e., only attending at fixed intervals. Models such as Sparse Transformer~\citep{child2019generating} and/or Longformer~\citep{beltagy2020longformer} employ strided or ``dilated'' windows.
    \item \textbf{Compressed Patterns} - Another line of attack here is to use some pooling operator to down-sample the sequence length to be a form of fixed pattern. For instance, Compressed Attention~\citep{liu2018generating} uses strided convolution to effectively reduce the sequence length.
\end{itemize}
\item \textbf{Combination of Patterns (CP)} - The key idea of combined\footnote{We note that this is also often referred to as factorization approaches, e.g., in~\citet{child2019generating}. We decide to refer to this class of models as combination approaches because (1) it is a better fit to what these models are actually doing and (2) to avoid confusion with matrix factorization or low-rank approaches. } approaches is to improve coverage by combining two or more distinct access patterns. For example, the Sparse Transformer~\citep{child2019generating} combines strided and local attention by assigning half of its heads to each pattern. Similarly, Axial Transformer~\citep{ho2019axial} applies a sequence of self-attention computations given a high dimensional tensor as input, each along a single axis of the input tensor. In essence, the combination of patterns reduces memory complexity in the same way that fixed patterns does. The difference, however, is that the aggregation and combinaton of multiple patterns improves the overall coverage of the self-attention mechanism.
\item \textbf{Learnable Patterns (LP)} - An extension to fixed, pre-determined pattern are \emph{learnable} ones. Unsurprisingly, models using learnable patterns aim to learn the access pattern in a data-driven fashion. A key characteristic of learning patterns is to determine a notion of token relevance and then assign tokens to buckets or clusters~\citep{vyas2020fast, wang2020cluster}. Notably, Reformer~\citep{kitaev2020reformer} introduces a hash-based similarity measure to efficiently cluster tokens into chunks. In a simlar vein, the Routing Transformer~\citep{roy2020efficient} employs online $k$-means clustering on the tokens. Meanwhile, the Sinkhorn Sorting Network~\citep{tay2020sparse} exposes the sparsity in attention weights by learning to to sort blocks of the input sequence. In all these models, the similarity function is trained end-to-end jointly with the rest of the network. The key idea of learnable patterns is still to exploit fixed patterns (chunked patterns). However, this class of methods learns to sort/cluster the input tokens - enabling a more optimal global view of the sequence while maintaining the efficiency benefits of fixed patterns approaches.
% to determine the optimal sparsity pattern.
\item \textbf{Neural Memory} - Another prominent method is to leverage a learnable side memory module that can access multiple tokens at once. A common form is \emph{global} neural\footnote{We use the term neural here to refer to a representation-like memory that is often manifested in the model.} memory which is able to access the entire sequence. The global tokens act as a form of model memory that learns to gather from input sequence tokens. This was first introduced in Set Transformers~\citep{lee2019set} as the \textit{inducing points} method. These parameters are often interpreted as ``memory'' and are used as a form of \textit{temporary} context for future processing. This can be thought of as a form of parameter attention~\citep{sukhbaatar2019augmenting}. Global memory tokens are also used in ETC~\citep{ainslie2020etc} and Longformer~\citep{beltagy2020longformer}. With a limited amount of neural memory (or inducing points), we are able to perform a preliminary \textit{pooling}-like operation of the input sequence to compress the input sequence - a neat trick to have at one's disposal when designing efficient self-attention modules.

\item \textbf{Low-Rank Methods} - Another emerging technique is to improve efficiency by leveraging low-rank approximations of the self-attention matrix. The key idea is to assume low-rank structure in the $N\times N$ matrix. The Linformer~\citep{wang2020linformer} is a classic example of this technique, as it projects the length dimension of keys and values to a lower-dimensional representation ($N \rightarrow k$). It is easy to see that the low-rank method ameliorates the memory complexity problem of self-attention because the $N \times N$ matrix is now decomposed to $N \times k$.
\item \textbf{Kernels} - Another recently popular method to improve the efficiency of Transformers is to view the attention mechanism through kernelization. The usage of kernels~\citep{katharopoulos2020transformers,choromanski2020masked} enable clever mathematical re-writing of the self-attention mechanism to avoid explicitly computing the $N \times N$ matrix. Since kernels are a form of approximation of the attention matrix, they can be also viewed as a type of low-rank approach~\citep{choromanski2020masked}. Examples of recent work in this area include Performers, Linear Transformers and Random Feature Attention (RFA,~\citep{peng2021random})
\item \textbf{Recurrence} - A natural extension to the blockwise method is to connect these blocks via recurrence. Transformer-XL~\citep{dai2019transformer} proposed a segment-level recurrence mechanism that connects multiple segments and blocks. These models can, in some sense, be viewed as \textit{fixed pattern} models. However, we decided to create its own category due to its deviation from other block / local approaches.
\item \textbf{Downsampling} - Another popular method of reducing computation cost is to reduce the resolution of the sequence, hence reducing computation costs by a commensurate factor. Examples of this class of models include Perceiver~\citep{jaegle2021perceiver}, Funnel Transformers~\citep{dai2020funnel}, Swin Transformer~\citep{liu2021swin}, and Charformer~\citep{tay2021charformer} models. Notably, there might also be some form of overlap of this class of models with models that leverage \textit{memory} tokens as models such as Set Transformer can also be viewed as a form of downsampling, albeit within the attention mechanism. The recent Nystr\"{o}mformer~\citep{xiong2021nystr}, on the surface, may seem like a low-rank or kernal-based approach. However, it is actually a downsampling approach where the \textit{`landmarks`} are simply strided based pooling - in similar spirit to Set Transformer, Funnel Transformer or Perceiever. 
\item \textbf{Sparse Models and Conditional Computation} - While not targeted specifically at the attention modules, sparse models \textit{sparsely} activate a subset of the parameters which generally improves the parameter to FLOPs ratio. Examples of this class of model includes Switch Transformers~\citep{fedus2021switch}, ST-MoE \citep{zoph2022designing}, GShard~\citep{lepikhin2020gshard}, Product-Key Memory Layers~\citep{lample2019large}. Within the scope of our studied models, sparse models typically operate on an adaptive basis in which the sparsity is typically learned (via mixture-of-experts like mechanism). Within this context, we can also consider sparsification of attention weights to fall under this paradigm. For this reason, we believe there is a close connection to fixed or learned patterns in attention. However, we believe that the emergence of an entire research direction~\citep{roller2021hash,lewis2021base,lepikhin2020gshard,du2021glam} based on sparse efficient should warrant a new category of efficient Transformers.

\end{itemize}
We note that these buckets are a broad characterization of the different efficient Transformer models. In reality, there is no sharp boundary between the buckets as models may be comprised of multiple technical innovations. For example, the $k$-means clustering in Routing Transformer~\citep{roy2020efficient} can also be interpreted as a form of global model memory approach, since one can view the centroids as parameterized model memory. In Reformer, however, clustering is used to learn the sparsity pattern of the attention weights. Additionally, pooling~\citep{liu2018generating} can be also interpreted as a form of model memory mechanism. We also note that the recent xformer models (circa December 2021) have started adopting some form of two-staged attention mechanism. Many times, these attention mechanisms explicitly combine one or more flavours of the above, e.g., local windows and then memory in Poolingformer~\citep{zhang2021poolingformer}, or Long Short Transformers~\citep{zhu2021long} that utilize low rank attention with fixed windows (e.g., a combination of local attention with Linformer-like inductive bias).

\subsection{Detailed Walk-through of Efficient Transformer Models}
This section delves into the details of several key efficient Transformer models, discussing their pros, cons, and unique talking points. The goal here is not to exhaustively detail all such models, but rather to cover a representative sample of models. 

\paragraph{Structure of this section} We begin by discussing local and fixed patterns models such as the Memory Compressed Transformer~\citep{liu2018generating} and Image Transformer~\citep{parmar2018image}. We then discuss the Set Transformers~\citep{lee2019set}, an early approach for utilizing global model memory. Following which, we move on to models that utilize combinations of patterns such as Sparse Transformers~\citep{child2019generating}, CCNet~\citep{huang2019ccnet}, and Axial Transformers~\citep{ho2019axial}. Next, we discuss Longformer~\citep{beltagy2020longformer} and ETC~\citep{ainslie2020etc}, as examples of memory-based Sparse Transformer  approaches. Our detailed walkthrough then moves on to models that incorporate learnable patterns (LP) such as Routing Transformers~\citep{roy2020efficient}, Reformer~\citep{kitaev2020reformer} and Sinkhorn Transformers~\citep{tay2020sparse}. After which, we introduce Linformer~\citep{wang2020linformer} and Synthesizers~\citep{tay2020synthesizer}, models that can be considered low-rank factorization approaches. We then discuss models based on kernel approaches such as Performer~\citep{choromanski2020masked} and Linear Transformers~\citep{katharopoulos2020transformers}. Following which, we discuss the models that are based on segment-based recurrence such as Transformer-XL~\citep{dai2019transformer} and Compressive Transformers~\citep{rae2020compressive}. Finally, we discuss the family of Sparse models which primarily leverage Mixture-of-Experts (MoE) type architectures and conditional computation to achieve computational efficiency. The logical flow of this section is aimed to be loosely chronological instead of categorically organized (with the exception of certain buckets like recurrence or sparsity that are more orthogonal approaches). We believe this is pedagogically helpful.

\subsubsection{Memory Compressed Transformer}
Memory Compressed Transformer~\citep{liu2018generating} is one of the early attempts at modifying Transformers to better handle longer sequences. The modification introduced by Memory Compressed Transformers is in two folds: localizing the attention span and using memory compressed attention.  

\paragraph{Local Attention Span}
A straightforward solution for dealing with long sequences in Transformers is to limit the attention span to a local neighborhood. \citet{liu2018generating} proposed dividing the input sequence into blocks of similar length so that self-attention can be computed within each block independently. This keeps the cost of attention per block constant, thus the number of activations scales linearly with the input length. 

\paragraph{Memory-compressed Attention}
The idea behind memory compressed attention is to reduce the number of keys and values using a strided convolution, while the queries remain unchanged. This leads to a reduction in the size of the attention matrix as well as the attention computations based on a compression factor that depends on the kernel size and the strides of the convolution. Memory compressed attention lets the model exchange the information globally across the input sequence as opposed to local attention. 

\paragraph{Computation and Memory Complexity}
For a block size of $b$, the computational and memory cost of self-attention in each block is $\mathcal{O}(b^2)$. Given there are $n/b$ blocks, the computational and memory cost of local attention is $\mathcal{O}(b.n)$. For memory-compressed attention, applying a convolution with kernel size and strides of $k$, the computational and memory cost of the attention mechanism reduces to $\mathcal{O}(n \cdot n/k)$. 

\subsubsection{Image Transformer}
Image Transformer~\citep{parmar2018image}, inspired by convolutional neural networks, restricts the receptive field of self-attention to only local neighborhoods. This helps the model scale up to process larger batch sizes while keeping the likelihood loss tractable. Besides the efficiency, adapting the notion of locality can be a desirable inductive bias for processing images. Image Transformer offers the encoder-decoder architecture, where the encoder generates a contextualized representation for every pixel-channel in the inputs and the decoder autoregressively generates one channel per pixel at each time step.

\paragraph{Localized Attention Span}
Limiting the receptive field to a local neighborhood~\citep{parmar2018image, parmar2019stand} addresses the issues with the computational and memory costs of running global self-attention on large inputs, but changing the neighborhood per query position would prohibit packing the computations of the self-attention into two matrix multiplications. To avoid that, Image Transformer proposes partitioning the inputs into ``query blocks’’ and their associated ``memory blocks``, where for all queries from a single query block, the model attends to the same memory block.
There are two different schemes for choosing query blocks and their associated memory block neighborhoods: \emph{1-dimensional local attention} and \emph{2-dimensional local attention}. Here we briefly explain these schemes in the decoder case. 

\begin{figure}
     \centering
     \begin{subfigure}[b]{0.4\textwidth}
         \centering
         \includegraphics[width=\textwidth]{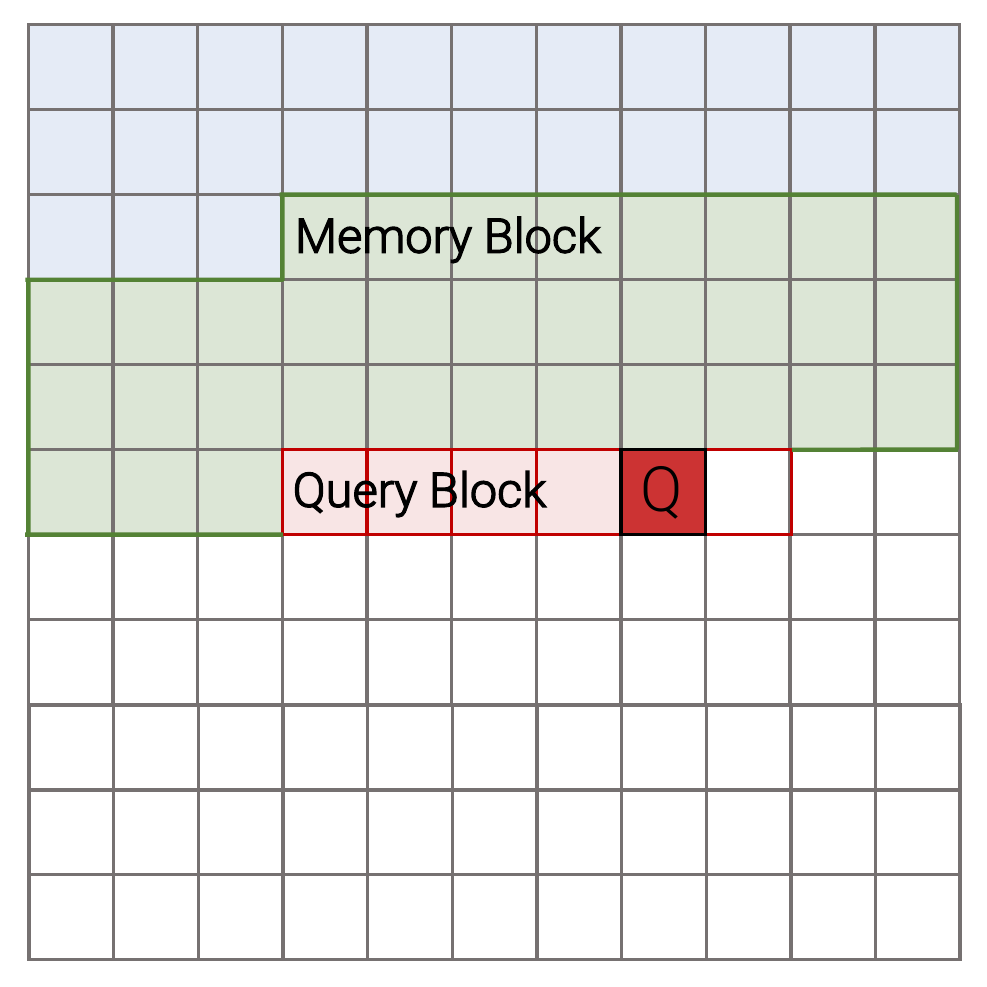}
         \caption{1-dimensional local attention}
         \label{fig:image_transformer-1d}
     \end{subfigure}
     \hfill
     \begin{subfigure}[b]{0.4\textwidth}
         \centering
         \includegraphics[width=\textwidth]{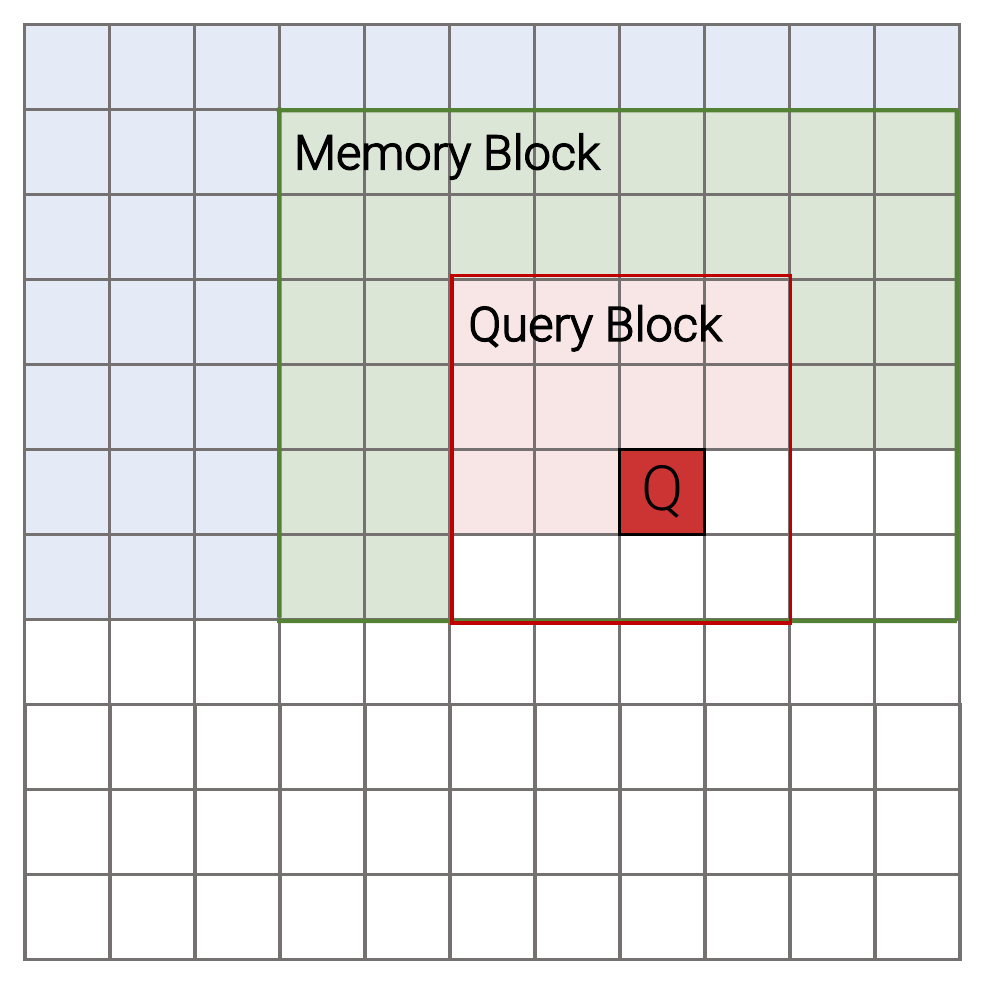}
         \caption{2-dimensional local attention}
         \label{fig:image_transformer-2d}
     \end{subfigure}
     \caption{Attention span in Image Transformer on a two-dimensional input.}
    \label{fig:image_transformer}
\end{figure}

For the 1-dimensional local attention, the image is flattened in the raster order\footnote{Given a 2D image as a grid of pixels, the horizontally left-to-right scanning of pixels, line-by-line, creates a raster order.} and partitioned into non-overlapping query blocks $Q$ of length $l_q$, and for each query block, a memory block $M$ is built from the same pixels in the $Q$ as well as a fixed number of pixels, $l_m$, generated before the query pixel.  In 2-dimensional local attention, pixels are generated in raster order. 
For the 2-dimensional local attention, the image is partitioned into multiple non-overlapping rectangular query blocks of length $l_q = w_q \times h_q$. The memory block extends the query block to the top, left $h_m$ and  $w_m$ pixels and to the right $w_m$ pixels, so $l_m = (w_q \times q_h) + 2 \times (h_m  + w_m)$.
The query pixel can attend to all other pixels. In the 2-dimensional local attention, pixels in the image are generated one query block after another. Generated blocks are in raster order, as well as generated pixels inside every block. 

\paragraph{Computational and Memory Complexity} 
In Image Transformer, the attention matrix has the shape of $l_q \times m$, where $l_q$ is the chosen length for the query blocks and $M$ is the length of the memory block (which is in fact $l_q + l_m$). Given that memory blocks do not overlap, we have to compute $n \times l_q$ attention matrices. Thus the memory and computational complexity of Image Transformer is $\mathcal{O}(n\cdot m)$.

\paragraph{Restrictions}
Image Transformer, and in general restricting the context in the attention mechanism to a local neighborhood, can decrease the cost of memory and computation at the price of losing the global receptive field. This can be an issue where global information is required to solve the task.  Also, local-attention has quadratic complexity with respect to the region length, thereby introducing an extra hyper-parameter in the trade-off between performance and computational complexity.

\subsubsection{Set Transformer}
The Set Transformer~\citep{lee2019set} adapts the Transformer model for \emph{set-input} problems - that is, problems wherein the input is a set of features and the output is some function of this set (and is thereby invariant to the permutation, or ordering, of the input features). The Set Transformer leverages attention to capture interactions between elements of the input set. Furthermore, it applies the idea of \emph{inducing points} from the sparse Gaussian process literature to reduce the complexity of attention from quadratic to linear in the size of the input set.

Problems involving sets of objects often have a \emph{permutation invariance} property: the target value for the set is the same regardless of the order of the objects in the set. \citet{zaheer2017deep} proved that all permutation-invariant functions can be represented by the following functional form:
\begin{align*}
\text{network}\left(\{x_1,\dots, x_N\}\right) = \rho\left(\text{pool}\left(\{\phi(x_1),\dots,\phi(x_N)\}\right)\right),
\end{align*}
where the pooling function $\text{pool}$ is a simple summation and $\phi$ and $\rho$ are continuous functions. This form can be interpreted as the composition of an \emph{encoder} $\phi$ and \emph{decoder} $\rho\left(\text{pool}(\cdot)\right)$.
While this form is a universal approximator in the space of permutation-invariant functions, it is unclear how well such models fit tasks in practice. The Set Transformer proposes a solution that can be viewed as an encoder and pooled decoder, but where, unlike the form given above, the encoder and decoder can attend to input elements individually and the pooling function is parameterized.
\paragraph{Attention Blocks}
The model introduces the following constructs: ``Multihead Attention Block'' (MAB), ``Set Attention Block'' (SAB), ``Induced Set Attention Block'' (ISAB), and ``Pooling by Multihead Attention'' (PMA). They are defined as follows.
\begin{align*}
\mathbf{\textbf{MAB}(X, Y)} &:= \text{LayerNorm}\left(H + \text{rFF}(H)\right),\\
H &:= \text{LayerNorm}\left(X + \text{MultiheadAttention}(X, Y)\right),\\
\mathbf{\textbf{SAB}(X)} &:= \text{MAB}(X, X),\\
\mathbf{\textbf{ISAB}_m(X)} &:= \text{MAB}\left(X, \text{MAB}(I_m, X)\right).\\
\mathbf{\textbf{PMA}_k(X)} &:= \text{MAB}\left(S_k, \text{rFF}(X)\right).
\end{align*}
Here, $X \in \reals^{N \times d}$ represents $N$ $d$-dimensional input/outputs stacked row-wise and $\text{rFF}$ is a parameterized feed-forward layer that operates on each row of its input matrix separately. $I_m \in \reals^{m \times d}$ represents $m$ \emph{trainable} $d$-dimensional ``inducing points'' while $S_k \in \reals^{k \times d}$ represent $k$ trainable $d$-dimensional ``seed vectors'' (with $k$ set to $1$ except when $k > 1$ correlated outputs are needed).
The Set Transformer's encoder is just $N$ layers of either SAB or ISAB (with $N$ often set to $2$ in practice) while its decoder is given by:
\begin{align*}
\mathbf{\textbf{Decoder}(X)} := \text{rFF}\left(\text{SAB}\left(\text{PMA}_k(X)\right)\right).
\end{align*}
It is straightforward to see that both ISAB and SAB are \emph{permutation equivariant} - in other words, if the input is permuted in some way then the corresponding output of the block is permuted in exactly the same way. Meanwhile, the pooling layer PMA is permutation invariant. Since functional composition, i.e. layering, preserves these properties, the Set Transformer encoder-decoder combination is permutation invariant.
\paragraph{Efficiency}
We can understand the $m$ inducing points $I_m$ learned in each ISAB layer as a form of static model memory. In addition to reducing the $\mathcal{O}(N n^2)$ complexity of the self-attending SAB layer to $\mathcal{O}(N m n)$, a reduction particularly valuable when the input set is large, the inducing points effectively encode some global structure that helps explain its inputs. For example, in the problem of \emph{amortized clustering}, where one attempts to learn to map an input set of points to the centers of clusters of points inside the set, the inducing points learned could be appropriately distributed so that the encoder can effectively compare query elements with each other implicitly via their proximity to the inducing points.

The trainable $k$ seeds $S_k$ used in the pooling layer $\text{PMA}_k$ can be viewed as static model memory in a similar light, reducing the memory and runtime complexity of the architecture. 

\begin{figure}
     \centering
     \begin{subfigure}[b]{0.4\textwidth}
         \centering
         \includegraphics[width=\textwidth]{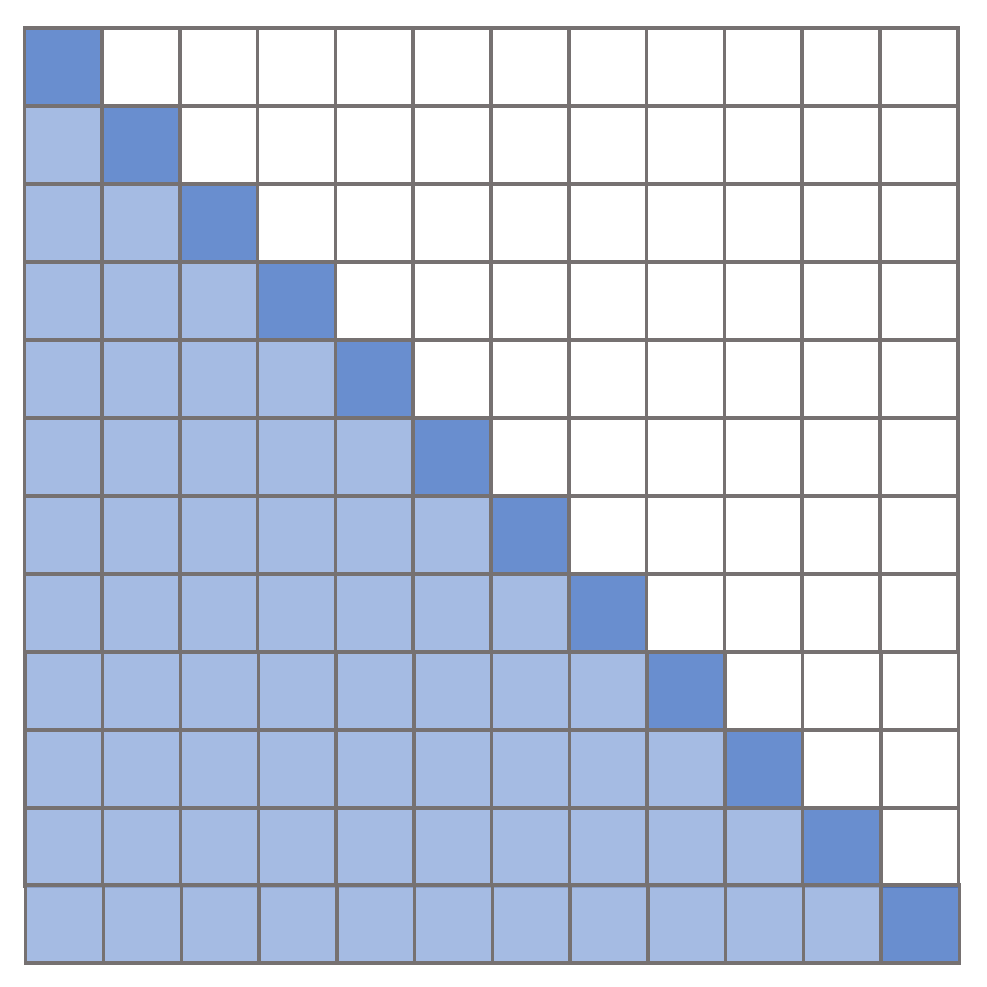}
         \caption{Transformer}
         \label{fig:dense_att}
     \end{subfigure}
     \hfill
     \begin{subfigure}[b]{0.4\textwidth}
         \centering
         \includegraphics[width=\textwidth]{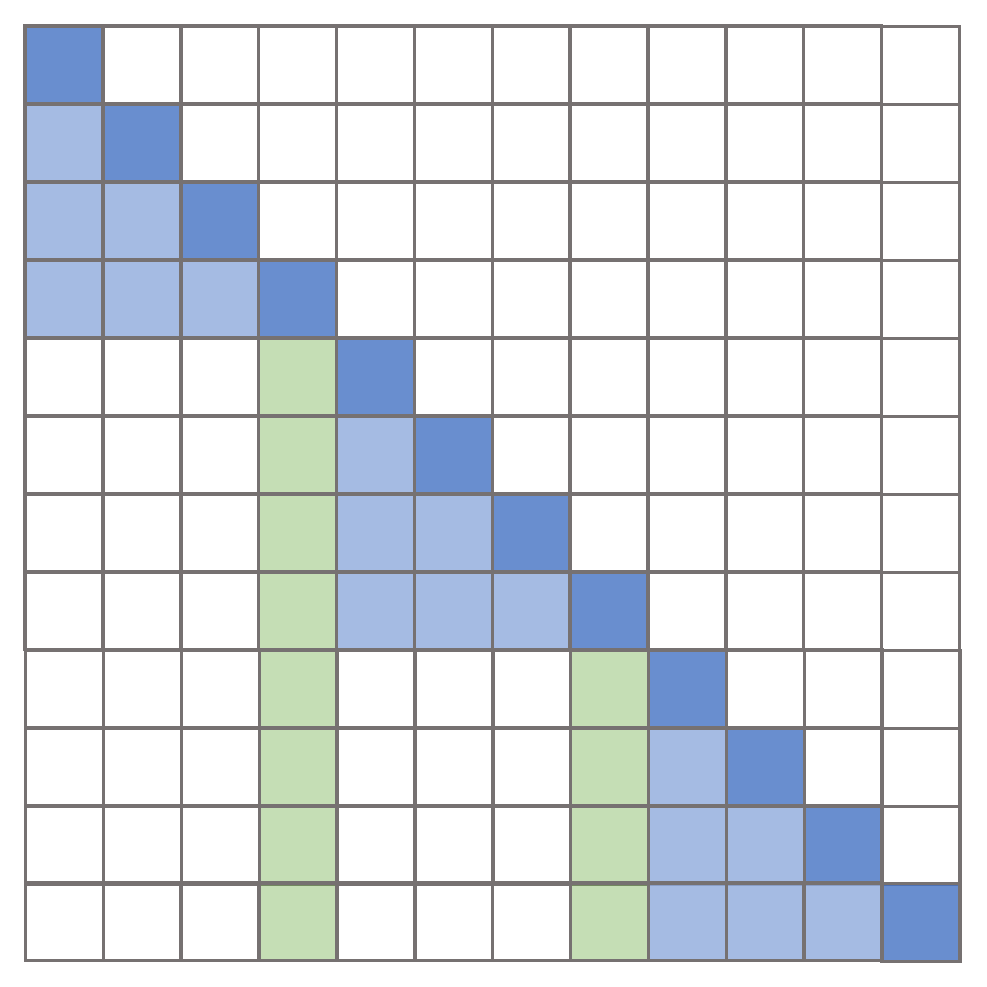}
         \caption{Sparse Transformer}
         \label{fig:sparse_att}
     \end{subfigure}
     \caption{Illustration of patterns of the attention matrix for dense self-attention in Transformers and sparse fixed attention in Sparse Transformers. Blue in the right diagram represents the local self-attention while green represents the strided component of the sparse attention.}
    \label{fig:sparse_transformer}
\end{figure}

\subsubsection{Sparse Transformer}
The Sparse Transformer~\citep{child2019generating} presents a simple initial attempt to reduce the quadratic complexity of the standard self-attention mechanism. The key idea is to reduce the dense attention matrix to a sparse version by only computing attention on a sparse number of $q_i,k_j$ pairs. Sparse Transformer employs fixed attention patterns which are defined by strides and local neighborhoods. Computation is \textit{factorized}, wherein local and stride patterns are split amongst the heads. 
\paragraph{Local Attention Heads} Half of the heads in the Sparse Transformer are dedicated to local attention.
\begin{align*}
    \hat{A}_{ij} = 
    \begin{cases}
    Q_{i}(K)_{j}^\top),& \text{if } \lfloor{{j}/{N}}\rfloor = \lfloor{i/{N}}\rfloor\\
    0              & \text{otherwise}
\end{cases}
\end{align*}
where $A_{ij}$ is the attention weight of $q_i,k_j$ and $\lfloor \: \rfloor$ denote the floor operation. In this case, we only compute the attention if $\lfloor{{j}/{N}}\rfloor = \lfloor{i/{N}}\rfloor$ (within the same block). 
\paragraph{Strided Attention Heads} The other half of the heads are dedicated to fixed strided patterns. Concretely,
\begin{align*}
    \hat{A}_{ij} = 
    \begin{cases}
    Q_{i}(K)_{j}^\top),& \text{if } (i-j) \mod N =0 \\
    0              & \text{otherwise}
\end{cases}
\end{align*}
The final result of the factorized sparse attention is visualized in Figure~\ref{fig:sparse_transformer}. We refer interested to~\citep{yun2020n} for some additional theoretical analysis about the expressiveness of the Sparse attention mechanism.
\paragraph{Parameter and Memory Complexity} The modification in the self-attention mechanism does not alter the parameter costs of the model since the model still retains the $Q,K,V$ transforms from the original Transformer model. The memory complexity of the attention layer is reduced from $\mathcal{O}(n^2)$ to $\mathcal{O}(n\log n)$ . 

\paragraph{Restrictions} The Sparse Transformer implementation requires custom GPU kernels to implement a specific block-sparse variant of matrix-matrix-multiplication and cannot be easily implemented on other hardware such as TPUs.

\subsubsection{Axial Transformer}

Axial Transformer~\citep{ho2019axial, weissenborn2019scaling} uses factorization in a simple yet effective setup for the self-attention mechanism to process large inputs that are organized as multidimensional tensors. Instead of applying attention to the flattened version of the input, Axial Transformer simply applies multiple attentions, each along a single axis of the input tensor. Each attention, in fact, mixes information along a particular axis, while keeping information along other axes independent.  Since the length of any single axis is typically much smaller than the total number of elements, Axial Transformer significantly saves computation and memory. 
\begin{figure}[t]
    \centering
    \includegraphics[width=0.4\linewidth]{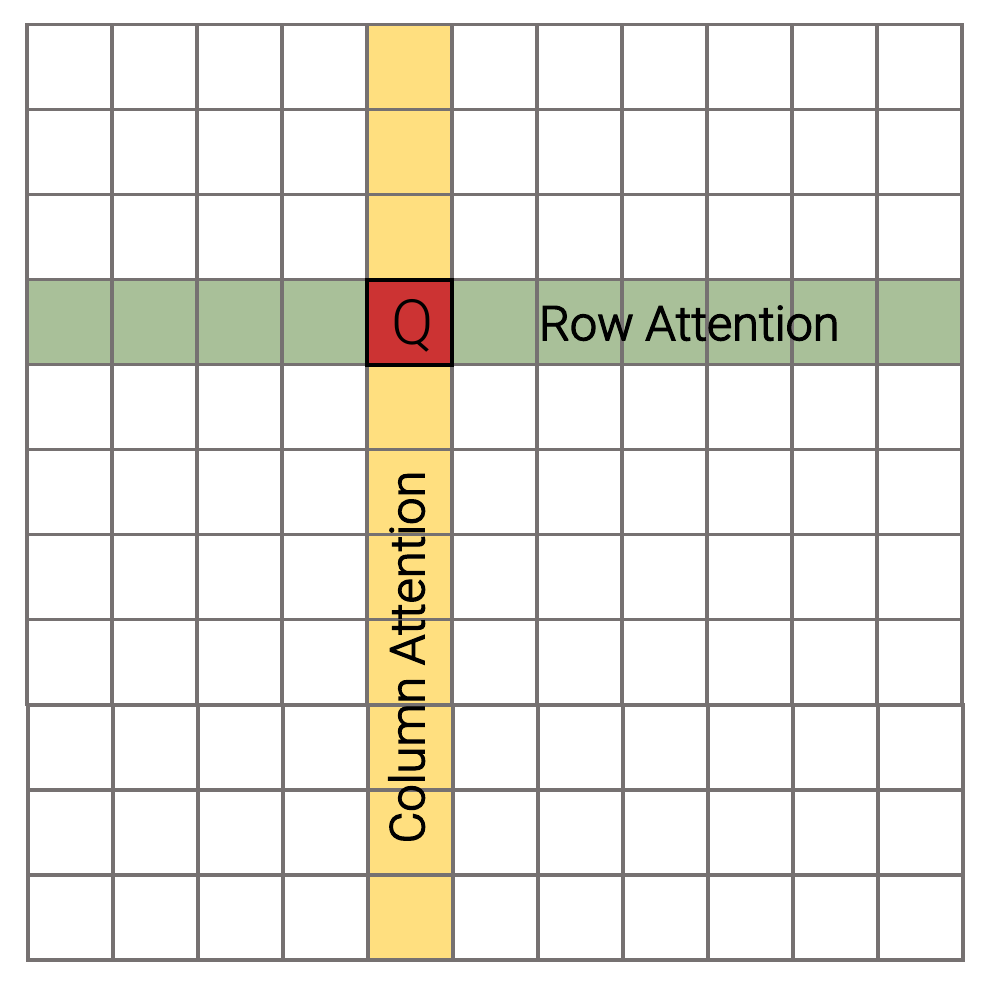}
    \caption{Attention span in Axial Transformer on a two-dimensional input.}
    \label{fig:axial_transforme}
\end{figure}

Axial Transformer offers an encoder-decoder architecture. For the decoding, to be able to implement the causal mask, Axial Transformer combines axial attentions with shift operations. For instance, for a model on 2-dimensional tensors, pixels are generated in raster order and to do that, first, the model encodes all pixels through an unmasked row and unmasked column attention. Then, for each row, the model applies an unmasked row and masked column attention to integrate the previously sampled rows. Finally, the model shifts the encoded representation up to make sure the conditioning information satisfies causality, and runs a masked row-attention to sample a new row in the image.

An advantage of Axial Transformer over similar methods like Sparse Transformer is that while it provides the global receptive field, it is straightforward to implement and does not require a custom kernel for an efficient implementation.

\paragraph{Computational and Memory Complexity} 
In terms of memory and computational complexity, on a square image of size $N$, Axial Transformer performs the attention computation in $\mathcal{O}(n \sqrt{n})$, which saves   $\mathcal{O}(\sqrt{n})$ over normal self-attention. For instance, with on square image with $N$ pixels, organized in a $b\times b$ grid, Axial Transformer runs $b$ attention sequences of length $b$,  which is of complexity $\mathcal{O}(b.b^2)$. In a more general case, for a $d$-dimensional tensor of shape $N = N^{1/d}\times \ldots \times N^{1/d }$, Axial Transformer saves a  $\mathcal{O}(N^{(d-1)/d})$ factor of resources over standard self-attention.

\subsubsection{Longformer}
Longformer~\citep{beltagy2020longformer} is a variant of Sparse Transformer. 
% Fundamentally, the Longformer is very similar to the Sparse Transformer.
Its key distinction compared to Sparse Transformer is ``Dilated Sliding Windows'', which can enable better long-range coverage without sacrificing sparsity. This is achieved by increasing the receptive fields by having gaps in the attention patterns. The Longformer also gradually increases the receptive field as the model goes deeper, dedicating lower levels for modeling local patterns and upper levels for modeling global patterns. 
\paragraph{Global Attention} For classification tasks, Longformer adopts global memory tokens that have access to all input sequences.
\paragraph{Parameter and Memory Complexity} The complexity of the model is reduced from $\mathcal{O}(n^2)$ to $\mathcal{O}(nk)$ where $k$ is the size of the window. When using global attention, the Longformer creates another set of query-key-value
% $QKV$
projections for this global attention, doubling the cost of the parameters at the attention layer. 

\subsubsection{Extended Transformer Construction (ETC)} 
The ETC model~\citep{ainslie2020etc} is another variation in the Sparse Transformer family. It introduces a new global-local attention mechanism. There are four components to this new attention mechanism, namely (1) global-to-global (g2g), global-to-local (g2l), local-to-global (l2g) and local-to-local (l2l). Aside from the original input to the model, ETC introduces $n_g$ auxiliary tokens as a prefix to the original input sequence. These tokens are regarded as global tokens and take part in global-to-$*$ and $*$-to-global attention. The local-to-local component acts as the local attention with a fixed radius of $k$. Overall, ETC is quite similar to Longformer in the way it introduces global auxiliary tokens. These tokens are trainable parameters and can be interpreted as a form of model memory that pools across the sequence to collect global sequence information. 

\paragraph{Memory and Parameter Complexity} The memory complexity of the ETC model is $\mathcal{O}(n_{g}^2 + n_{g}N)$, where $n_g$ is the number of global tokens and $N$ is the input sequence length.

\paragraph{Restrictions} Intuitively, it is easy to observe that ETC cannot be used for auto-regressive decoding. This is because we are not able to compute causal masks because of the global attention. 

\subsubsection{BigBird}
The BigBird model~\citep{zaheer2020big} is another Transformer for modeling longer sequences and is primarily built on top of ETC~\citep{ainslie2020etc}. The Big Bird model is comprised of several key components, namely (1) global tokens, (2) random attention (queries attend to random keys), and (3) fixed patterns (local sliding windows). 

\paragraph{Global Attention} Fundamentally, the idea of using global model memory can be traced all the way back to Longformer/ETC and Set Transformer model.  Notably, the global model memory in Big Bird is extended to contain tokens within the sequence, instead of simply parameterized model memory. The authors call this the \textit{`internal transformer construction (ITC)'} in which a subset of indices is selected as global tokens. This can be interpreted as a model-memory-based approach. 

\paragraph{Sliding Window Attention} The window-ed attention was first proposed in early local-based attention models (Image Transformer, Compressed Attention and/or Sparse Transformer). In BigBird, each query attends to $w/2$ tokens to the left and $w/2$ tokens to the right. This corresponds to a fixed pattern (FP) approach.

\paragraph{Random Attention} Finally, each query attends to $r$ random keys. This pattern is fixed. 

\paragraph{Memory and Parameter Complexity} The memory complexity of the self-attention is linear, i.e., $\mathcal{O}(n)$. The BigBird model does not introduce new parameters beyond the Transformer model. 

\paragraph{Restrictions} Similar to ETC, the BigBird model cannot be used to autoregressively decode. Hence, qualifying it as an encoder-only model.

\subsubsection{Routing Transformer}
The Routing Transformer~\citep{roy2020efficient} is a content-based sparse attention mechanism. It proposes a clustering-based attention mechanism that learns the attention sparsity in a data driven fashion. The first step is to project $Q$ and $K$ into a routing matrix $R$ of dimensions $n \times d$.
\begin{align}
R = QW_R + KW_R     
\end{align}
where $W_R$ is a $d \times d$ orthonormal projection matrix. 

\paragraph{$k$-means Clustering} The $R$ matrix undergoes $k$-means clustering with a series of parameterized cluster centroids $u_1, u_2 \cdots c_k$. The $k$-means in Routing Transformer is trained in an online fashion. To ensure a similar number of tokens in each cluster, the model initializes $\sqrt{n}$ clusters, computes each token's distance against the cluster centroid, and takes an equal top-$k$ for each centroid. Since the cluster centroids are trainable parameters, this is also reminiscent of the \emph{all-attention} layer proposed by~\citep{sukhbaatar2019augmenting}. 

\paragraph{Routing Strategy} The routing strategy is then defined as:
\begin{align}
X'_i = \sum_{j \in C_i, j \leq i} A_{ij} V_j    
\end{align}
where $C_i$ is the cluster that vector $R_i$ is assigned to. In other words, the token at $i$ only attends to tokens in the same cluster.

\paragraph{Memory and Parameter Complexity} The Routing Transformer introduces additional parameters in the clustering mechanism, namely $k \times d$ centroid vectors and a $W_r$ projection matrix. The memory complexity is $\mathcal{O}(n^{1.5})$. 

\subsubsection{Reformer}
Reformer~\citep{kitaev2020reformer} is another efficient attention model based on locality sensitive hashing (LSH). Reformer also introduces \emph{reversible} Transformer layers, which contribute to further reducing its memory footprint.

\paragraph{LSH Attention} The LSH attention introduces parameter-sharing between query and keys. It hashes the query-keys into buckets using a random-projection based hashing function. The key idea is that nearby vectors should obtain a similar hash while distant vectors should not, hence being termed as \textit{`locality sensitive'}. To perform hashing, a random matrix $R \in \reals^{k \times b/2}$ is first introduced. Next, The hashing function is defined as:
\begin{align}
h(x) = \text{arg max}([xR;-xR])    
\end{align}
where $[;]$ is the concatenation of two vectors. For all queries, attention is computed if and only if the query and key hashes match, i.e., $h(q_i)=h(k_j)$. In other words, attention is computed amongst query and keys if they fall in the same hash bucket. In order to maintain causal masking, Reformer assigns and maintains a position index for every query and key. It is therefore able to compare if each query key comparison is auto-regressively valid. 
\paragraph{Memory Efficiency with LSH Attention} The key idea behind LSH attention is to classify tokens into buckets and then process them bucket by bucket in a chunked fashion. To this end, queries are first sorted by bucket number and then by sequence order within the same bucket. During computation, tokens only attend to the same bucket in its own chunk and previous chunk. The chunking and sorted bucketing techniques help to improve the overall efficiency of the Reformer model.
% \paragraph{Reversible Layers} The Reformer introduces yet another efficiency trick for reducing memory consumption during training. Reversible layers~\citep{gomez2017reversible} reduce memory cost by enabling activations to be reconstructed from the next layer's. This reduces memory cost since this eliminates the need to store activations for all layers during backpropagation. RevNets operate on paired input/outputs, e.g., $(x_1,x_2) \rightarrow (y_1,y_2)$ mapping instead of a traditional single input $x \rightarrow y$ mapping. With $y_1 = x_1 + F(x_2)$ and $y_2 = x_2 + G(y_1)$, the layer is reversed via $x_2=y_2-G(y_1)$ and $x_1=y_1-F(x_2)$. In reversible Transformers, $F$ and $G$ are the attention function and the feed-forward function respectively. For the sake of brevity, we refer interested readers to~\citep{gomez2017reversible,kitaev2020reformer}.

\paragraph{Parameter and Memory Complexity} The memory complexity of Reformer is $\mathcal{O}(n \log n)$. In terms of parameter costs, Reformer shares queries and keys, which reduces the cost of the QKV transforms by a third. The random projections are not trainable parameters and hence do not incur parameter costs. Overall, Reformer has fewer parameters than vanilla Transformers. The reversible layers in Reformer also reduce the memory consumption during training by enabling activations to be reconstructed from the next layer's. This reduces memory cost since this eliminates the need to store activations for all layers during backpropagation.

\subsubsection{Sinkhorn Transformers}
This section introduces the Sparse Sinkhorn Transformer~\citep{tay2020sparse}. The Sinkhorn Transformer belongs to the family of \textit{learned patterns}. This model is a chunked/blocked model that learns sparse patterns by re-sorting the input key and values in a block-wise fashion and then applying local block-based attention. 
\begin{align*}
    A_{ij} = 
    \begin{cases}
    (Q_{i}\psi_S(K)_{j}^\top),& \text{if} \lfloor{{j}/{N}}\rfloor = \lfloor{i/{N}}\rfloor\\
    0              & \text{otherwise}
\end{cases}
\end{align*}
where $\psi_S$ applies a sorting operator on the sequence length dimension.  

\paragraph{Sorting Network} The sorting operator is parameterized by a meta sorting network. Let $X$ be the input sequence of dimension $N \times d$. 
\begin{equation}
\psi_S(X) = \phi_S(F_S(\textsc{BlockSum}(X)))\:\textsc{BlockShape}(X)
\end{equation}
where $F_S(.)$ is a parameterized function such as a two layer feed-forward network with ReLU activation. The output of $F_S(.)$ is a tensor of $n_B \times n_B$. The BlockSum function learns the sum embeddings of local blocks. The BlockShape function reshapes the input tensor into $\mathbb{R}^{N \times d} \rightarrow \mathbb{R}^{n_B \times b \times d}$. Here, we note that $N = n_B \times b$, where $b$ is the size of the block and $n_B$ is the number of total blocks.

\paragraph{Sinkhorn Sorting} $\phi$ is the Sinkhorn balancing operator~\citep{sinkhorn1964relationship, adams2011ranking} which converts the $n_B \times n_B$ matrix into a soft permutation matrix. Specifically, a series of row- and column-wise normalizations are applied on the matrix output of $F_S\text{BlockSum}(X)$. For the sake of brevity, we do not delve into details of this operation. Further details can be found at~\citet{adams2011ranking,tay2020sparse}.
\paragraph{Parameter and Memory Complexity} The memory complexity of the Sinkhorn Transformer is $\mathcal{O}(b^2)$ where $b$ is the block size and $b=\frac{N}{N_b}$. Additional parameter costs are incurred from the meta sorting network $F_S(.)$. The number of additional parameters is therefore $2d^2$ when a two layer ReLU network is used as the sorting network.

\subsubsection{Linformer}
Linformer~\citep{wang2020linformer} is an efficient Transformer based on the idea of low-rank self-attention. 
\paragraph{Low-Rank Projections on Length Dimensions} Linformer projects the $N \times d$ dimensional keys and values to $k \times d$ dimensions using additional projection layers. Note that this is a reduction on the length dimension instead of the key and value dimensions. This can 
Given the newly projected keys ($K'$) and values ($V'$), the $QK'$ matrix is now $(N \times k)$ dimensions instead of $(N \times N)$. The attention matrix $\text{Softmax}(QK')$ multiplies with $V' \in \mathbb{R}^{k \times d}$ to result in an output tensor of dimensions $N \times d$. To some extent, Linformer is reminiscent of depth-wise convolutions~\citep{kaiser2017depthwise}. A projection on the length dimension causes mixing of sequence information (dimension-wise) in a single transformation. Hence, it is non-trivial to maintain causal masking and/or prevent mixing of past and future information when computing attention scores. The formulation of Linformer (for each attention head) can be expressed as:
\begin{align}
Softmax(\frac{1}{\sqrt{d_k}}XW^{Q}_{i}(E_i X W_i^K)) \cdot F_iXW_i^V    
\end{align}
where $W^{Q,K,V}$ are the default linear transformation of $X$ into queries (as per vanilla Transformer) and $E_{i}, F_i$ are additional $k \times N$ projection of the key and values into $k \times d$ tensors.

\paragraph{Parameter and Memory Complexity} The memory complexity of Linformer is $\mathcal{O}(n)$. There is only a minimal parameter costs of the Linformer due to the extra $N \times k$ length projections. If $k$ is sufficiently small, there is negligible parameter costs incurred.

\subsubsection{Performer}
The Performer~\citep{choromanski2020masked, choromanski2020rethinking} model is characterized by its Generalized Attention mechanism and its usage of random Kernels. 
\paragraph{Generalized Attention}  The generalized attention entangles $Q_i,K_j$ with a kernel function $K$. The attention matrix in Performer is computed via:
\begin{align}
A = [g(Q_i^\top)K(Q_i^\top K_j^\top) h(K_j^\top)]    
\end{align}
where $K(.)$ is a kernel function that maps $d \times d$ to a scalar value $\mathbb{R}$ and $g,h$ are functions that map $d$ to a scalar value $\mathbb{R}$. 
\paragraph{Fast Attention via Orthogonal Random Features (FAVOR)} The above computation is still quadratic in complexity. Hence, the Performer leverages approximation tricks to avoid storing and computing the $N \times N$ attention matrix. It leverages \textit{orthogonal random features} (ORF) for doing so. The final attention output $Y$ of the Performer is described as follows:
\begin{align}
Y = \hat{D}^{-1}(Q'((K')^\top V))
\end{align}
where $\hat{D}=\text{diag}(Q'((K')^\top1_N))$, $Q'=D_Q\phi(Q^\top)^\top$, and $K'=D_K\phi(K^\top)^\top$. Note that $D_Q=g(Q_i^\top),D_K=h(K_i^\top)$. The function $\phi(x)$ is defined as:
\begin{align}
\phi(X)= \frac{c}{\sqrt{M}}f(Wx +b)^\top
\end{align}
where $c > 0$ is a constant, $W \in \mathbb{R}^{M \times d}$ is a random feature matrix and $M$ is the dimensionality of this matrix that controls the number of random features. We are able to see that we do not explicitly compute $A=QK^\top$ and hence avoid paying the $N^2$ cost. For rigorous theoretical analysis and further details, we refer interested readers to~\citep{choromanski2020masked}.
 
\paragraph{Parameter/Memory Complexity and Compute Costs} The complexity of the bi-directional FAVOR algorithm is $\mathcal{O}(Md + N d + MN)$ where $M$ is the dimensionality of the random features. It is worth noting that the unidirectional variations cannot be causally masked in an efficient linear-time fashion. As such, during training, running unidirectional (causal) implementation of kernel-based attention on an autoregressive task can be several times slower than vanilla Transformer during \textit{parallelized} training due to the need to do a left to right pass (i.e., scan operation) in similar spirit to Recurrent neural networks. Since many autoregressive tasks trained via parallelization and teacher forcing, this makes training Performer on a generative task prohibitively slow. In order for KV to be causally masked efficiently, one would have to manifest the $d \times d$ KV matrix at every time step - recovering a quadratic complexity model. We feel this is one of the intricate points that highlight how efficient memory complexity might not equate a faster or more efficient model in practice. We highlight that this only happens during autoregressive training. The inference-time for incremental decoding, however, would benefit from a speed up.

\subsubsection{Linear Transformer}
The Linear Transformer~\citep{katharopoulos2020transformers} improves the complexity of self-attention from quadratic to linear by using a kernel-based formulation of self-attention and the associative property of matrix products. Furthermore, it reduces attention with causal masking (which is used in auto-regressive decoding) to a linear-time, constant memory recurrent neural network (RNN). The model has been shown to improve inference speeds up to \emph{three orders of magnitude} without much loss in predictive performance. Linear Transformers are similar to Performers with the exception of the kernel function and therefore also suffer from the same drawbacks (unable to be parallelized across the time dimension during training in an autoregressive teacher forced setting). 

The method rests on the simple but powerful observation that the accumulated value $V_i'$ for the query $Q_i$ in position $i$ can be written as:
\begin{align*}
V_i' &= \frac{\sum_{j=1}^p \text{sim}(Q_i, K_j) V_j}{\sum_{j=1}^p \text{sim}(Q_i, K_j)}.
\end{align*}
Here, $p = N$ in full, unmasked attention and $p = i$ in the case of causal masking. Now, in usual softmax attention, $\text{sim}(q, k) = \exp\left(\frac{q^T k}{\sqrt{d}}\right)$. Linear Transformer, however, expresses the similarity as a kernel function. That is, $\text{sim}(q, k) := \phi(q)^T \phi(k)$, where $\phi$ is a, possibly high-dimensional, feature map. With this choice,
we can rewrite $V_i'$ as:
\begin{align*}
V_i' &= \frac{\phi(Q_i)^T S_p}{\phi(Q_i)^T Z_p},\\
S_p &:= \sum_{j=1}^p \phi(K_j) V_j^T,\\
Z_p &:= \sum_{j=1}^p \phi(K_j).
\end{align*}
For unmasked attention, since $p = N$ we only need to compute $S_N$ and $Z_N$ once and we reuse them for the computation at every position $0 \leq i \leq N$. For causal attention, the $S_i$'s and $Z_i$'s can be viewed as states of an RNN that are updated by the following recurrence relations:
\begin{align*}
S_i &= S_{i-1} + \phi(K_i)V_i^T,\\
Z_i &= Z_{i-1} + \phi(K_i)
\end{align*}
with initial condition $S_0 = Z_0 = 0$.
If the dimension of the key, query, and values are all $d$ and the cost to compute $\phi$ is $\mathcal{O}(c)$, then the overall run-time complexity of Linear Transformer is $\mathcal{O}{(N c d)}$. The authors choose
\begin{align*}
\phi(x) = \text{elu}(x) + 1,
\end{align*}
where $\text{elu}(\cdot)$ denotes the exponential linear unit~\citep{clevert2015fast}. With this choice of feature map, $c = d$ and the end-to-end complexity of the model is $\mathcal{O}(N d^2)$.

\subsubsection{Synthesizers}
Synthesizer models~\citep{tay2020synthesizer} are an attempt to study and investigate the true importance of conditioning within the self-attention mechanism and are also the first attempts at unconditional token-mixing. In~\citet{tay2020synthesizer}, the authors study a synthetic self-attention module in which attention weights are approximated instead of being computed by pairwise dot products. Synthesizers are only implicitly related to efficient Transformers and can be considered more as a MLP-Mixer~\citep{tolstikhin2021mlp}. However, the factorized variants can be considered a low-rank efficient Transformer model. 
\paragraph{Dense Synthesizers} In the Dense Synthesizer, each token $x_i$ is projected to a vector of length $N$ using a two-layered non-linear feed-forward network. The computation of the attention matrix $A$ is described as:
\begin{align}
A = W_2(\sigma_{R}(W_1(X)+b))+b    
\end{align}
where $X \in \mathbb{R}^{N \times d}$ is the input sequence, $W_2 \in \mathbb{R}^{d \times N}, W_1 \in \mathbb{R}^{d \times d}$, and $\sigma_R$ is the ReLU activation function. Given $A$, the output of the Synthetic Dense function is computed as:
\begin{align}
Y = \text{Softmax}(A)G(X).    
\end{align}
where $G(X)$ is another parameterized function $\mathbb{R}^{N \times d} \rightarrow \mathbb{R}^{N \times d}$. 
\paragraph{Random Synthesizers} Another variant of the Synthesizer model uses random matrices for $A$. In this case, the output can be expressed by:
\begin{align}
Y = \text{Softmax}(R)G(X).    
\end{align}
where $R \in \mathbb{R}^{N \times N}$ is a trainable and/or non-trainable matrix. In~\citet{tay2020synthesizer}, the authors show that Random Synthesizers achieve competitive performance.

\paragraph{Factorized Variants} The Dense and Random Synthesizers also come with factorized variants that consider a low-rank structure of the attention matrix. For factorized random Synthesizer can be written as:
\begin{align}
Y = \text{Softmax}(R_{1}R_{2}^{\top})G(X).    
\end{align}
where $R_{1},R_{2} \in \mathbb{R}^{N \times k}$. On the other hand, the Dense Synthesizer can be factorized as follows:
\begin{align}
A=H_B(B)* H_C(C)  \:\: \text{where} \: \: B, C = F_B(X_i), F_C(X_i), 
\end{align}
where $F_B(.)$ projects onto $b$ dimensions and $F_C(.)$ projects $X_i$ onto $c$ dimensions with $c \times b=N$. $H_B,H_C$ are tile and repeat functions respectively.
\paragraph{Parameter and Memory Complexity} For Random Synthesizers that adopt a non-trainable $R$, there is no need to store $N^2$ activations at this layer. For the trainable Random Synthesizer, the memory complexity and parameter complexity remains as $N^2$. However, there is no need to compute $N^2$ dot products, reducing the computational costs significantly. The Factorized Random Synthesizers reduce the parameter costs to $2(N \times k)$.

\subsubsection{Transformer-XL} The Transformer-XL model~\citep{dai2019transformer} relies on segment-based recurrence. Segment-based recurrence can be considered an orthogonal approach to the other techniques discussed since it does not explicitly sparsify the dense self-attention matrix. Instead, it connects adjacent blocks with a recurrent mechanism. % they condition segments (chunks) on previous chunks. 

\paragraph{Segment Recurrence} The recurrent mechanism in Transformer-XL is described as:
\begin{align}
\tilde{\bm{h}}^{n-1}_{\tau+1} &= [\text{SG}(\bm{h}^{n-1}_{\tau}) \odot \bm{h}^{n-1}_{\tau +1}]   \\ 
q^{n}_{\tau+1}, k^{n}_{\tau+1}, v^{n}_{\tau+1} &= \bm{h}^{n-1}_{\tau+1}\bm{W}^\top_q \:,\: \tilde{\bm{h}}^{n-1}_{\tau+1}\bm{W}^\top_k \:,\: \tilde{\bm{h}}^{n-1}_{\tau+1}\bm{W}^\top_v \\ 
\bm{h}^{n}_{\tau+1} &= \text{Transformer}(q^{n}_{\tau+1}, k^{n}_{\tau+1}, v^{n}_{\tau+1})
\end{align}
where SG() is the stop gradient function, $\odot$ is the concatenation of two sequences along the length dimension. Notably, the keys and values are conditioned on the previous sequence length $\tilde{\bm{h}}^{n-1}_{\tau +1}$ instead of $\bm{h}^{n-1}_{\tau +1}$
\paragraph{Relative Positional Encodings} Transformer-XL introduces novel relative position encodings. In this scheme, absolute positional encodings are not added to the content embeddings. Instead, they are only considered while computing attention weights where they can be replaced with relative position encodings. Since the relative position encodings are not directly relevant to the efficiency of the model, we refer interested readers to~\citet{dai2019transformer} for more details. 

\subsubsection{Compressive Transformers}
Compressive Transformers~\citep{rae2020compressive} are a natural extension of the Transformer-XL model. The key idea behind the Compressive Transformer is to maintain a fine-grained memory of past segment activations. This is unlike Transformer-XL, which discards past activations as it moves across segments. 
\paragraph{Model Memory} The Compressive Transformer is characterized by a dual model memory system - a primary model memory and a secondary compressed model memory. It maintains a model memory with $n_m$ memory slots and $n_{cm}$ compressive memory slots. Whenever the model accepts a new input segment, the oldest $n_s$ activations in the primary model memory are moved to the compressed model memory where a compression function is applied. 

\paragraph{Compression} These memories are compressed with a variety of compression functions such as (1) mean/max pooling (2) 1D convolutions, (3) dilated convolutions, and (4) most used (e.g., sorted by usage of attention).  
\paragraph{Memory Reconstruction} In order to better retain memories over long sequences, the Compressive Transformer implements an auto-encoding loss that learns to reconstruct the original memory from its compressed version, i.e., $L^{ae}=|| \text{old\_mem} - g(\text{new\_cm}^{(i)})||$ where $g(.) : \mathbb{R}^{\frac{n_s}{c} \times d} \rightarrow \mathbb{R}^{n_s \times d}$ is a parameterized function. A second attention reconstruction is a lossy re-construct that attempts to reconstruct the attention over model memory instead of the lossless reconstruction of the model memory itself. 

% \begin{align}
% A_{ij} &= E^\top_{x_i} \bm{W}_q^\top W_{k,E} E_{x_j} + E^\top_{x_i}\bm{W}_q^\top \bm{W}_{k,R} R_{i-j} \\ &+ u^\top \bm{W}_{k,E} E_{x_j} + v^\top W_{k,R} R_{i-j}    
% \end{align}

\subsubsection{Sparse Models}
In this section we describe the family of Sparse models. Sparse models typically achieve a high parameter to FLOP ratio by sparsely activating a subset of parameters or activations. It is good to note that while most of the works within the scope of this survey deals with efficient attention, the scope of sparse models goes beyond the attention module and is generally applied more frequently to the feed forward layers~\citep{lepikhin2020gshard,fedus2021switch}. In this section, we discuss the prime variant for Sparse models, i.e., the Mixture-of-Experts based Sparse models which includes models such as GShard~\citep{lepikhin2020gshard}, Switch Transformer~\citep{fedus2021switch} and GLaM~\citep{du2021glam}.
\paragraph{Mixture-of-Experts} The key idea behind MoE is to route token $x_{i}$ to a set of selected experts determined by a routing function. The routing function typically computed a linear combination over experts using the softmax function and can be interpreted as a form of gating mechanism. The top-k gate values are then selected for each token $x_{i}$ and the final output of that layer is determined by a linear combination of selected top-k experts. This MoE layer remains foundational and fundamental to many MoE architectures, with the exception of certain implementation details. For example, Switch uses a top-1 routing strategy while GShard uses a group-level top-2 gating.

\section{Discussion}
This section explores the state of research pertaining to this class of efficient models. 
\subsection{On Evaluation}
While the field is bustling with new Transformer models, there is not an easy way to compare these models side by side. Many research papers select their own benchmarks to showcase the abilities of the proposed model. This is also coupled with different hyperparameter settings like model sizes and configurations which can make it difficult to correctly attribute the reason for the performance gains.
% with the modeling advance proposed in the paper itself.
Moreover, some papers conflate this with pretraining~\citep{devlin2018bert} which makes it even harder to distinguish the relative performance of these different models. It is still a mystery to which fundamental efficient Transformer block one should consider using.

On one hand, there are multiple models that focus on generative modeling, showcasing the ability of the proposed Transformer unit on auto-regressive modeling of sequences. To this end, Sparse Transformers~\citep{child2019generating}, Adaptive Transformers~\citep{correia2019adaptively}, Routing Transformers~\citep{roy2020efficient} and Reformers~\citep{kitaev2020reformer} are mainly focused on generative modeling tasks. These benchmarks typically involve language modeling and/or pixel-wise image generation on datasets such as wikitext~\citep{merity2016pointer}, and/or ImageNet~\citep{deng2009imagenet} / CIFAR~\citep{krizhevsky2009learning}. Models that use segment based recurrence such as Transformer-XL and Compressive Transformers are also focused on long-range language modeling tasks such as PG-19. 

On one hand, a collection of models is mainly focused on encoding-only tasks such as question answering, reading comprehension and or selections from the GLUE benchmark. For example, the ETC model~\citep{ainslie2020etc} only runs experiments on question answering benchmarks such as NaturalQuestions~\citep{47761} or TriviaQA~\citep{JoshiTriviaQA2017}. On the other hand, the Linformer~\citep{wang2020linformer} focuses on subsets of the GLUE~\citep{wang-etal-2018-glue} benchmark. This split is very natural and intuitive, since models like ETC and Linformer cannot be used in an auto-regressive fashion. This exacerbates the challenges associated with comparing these encoder-only models with the other models.

There are models that focus on a balance of both. Longformer~\citep{beltagy2020longformer} tries to balance this by running benchmarks on both generative modeling and encoder-only tasks. The Sinkhorn Transformer~\citep{tay2020sparse} compares on both generative modeling tasks as well as encoding only tasks. 

Additionally, it is also worth noting that, although Seq2Seq machine translation (MT) was one of the problems that popularized Transformer models, not many of these efficient Transformer models are evaluated on MT tasks. This is likely because sequence lengths in MT are not long enough to warrant the usage of these models. 

While generative modeling, GLUE tasks and/or question answering appear to be the common evaluation benchmarks adopted by many of these tasks, there are several niche benchmarks that a small isolated number of papers choose to evaluate on. For starters, the Performer model~\citep{choromanski2020masked} evaluates on masked language modeling on proteins, deviating from serious head-on comparisons with other efficient Transformer models. The Linear Transformer~\citep{katharopoulos2020transformers} also evaluates on speech recognition, which is a rare benchmark amongst this group of papers. 

There have been recent attempts to unify evaluation on Efficient Transformers, namely Long Range Arena, i.e., LRA, ~\citep{tay2020long} that benchmarked 10 different xformer variants on long range modeling tasks. It is good to note that LRA was designed for evaluating Transformers in encoder-only mode and do not consider generative (or autoregressive tasks) that require causal masking.

% Conclude w/ need for standardized benchmark?

\subsection{On Model Design Trends}
When matching our broad categorization against the timeline of the introduction of these models, we are able to see the trend that the community is taking towards designing efficient Transformer models. Early work in this area has primarilyy been focused on more intuitive and simple approaches such as \textit{fixed patterns}. To this end, most early work in this area is based on block/local patterns such as Image Transformer~\citep{parmar2018image}, Compressed Attention~\citep{liu2018generating}, Blockwise Transformer~\citep{qiu2019blockwise} or the local windows in Sparse Transformer~\citep{child2019generating}. 

The paradigm of factorizing various fixed patterns was first introduced in~\citet{child2019generating} and CCNet~\citep{huang2019ccnet}. Around this same time, we start to observe early traces of \textit{model-memory}-based approaches from both the inducing point methods in the Set Transformer~\citep{lee2019set} or global nodes in the Star Transformer~\citep{guo2019star} model.

We observe the next wave of models comes in the form of learnable sparsity patterns. Reformer~\citep{kitaev2020reformer} and Routing Transformers~\citep{roy2020efficient} are very similar in the sense that they are models that learn to cluster/bucket tokens before performing attention. The key difference is the means to the end whereby Reformer uses a hashing function while the Routing Transformer uses online $k$-means for cluster assignment. In parallel, Sinkhorn Transformers~\citep{tay2020sparse} are also based on the idea of sorting, albeit at the block level. These three models largely follow a similar paradigm of re-arranging sequences for efficient computation of attention scores.

Next, we then observe several extensions that are largely built off the Sparse Transformer paradigm. The ETC~\citep{ainslie2020etc} and Longformer~\citep{beltagy2020longformer} models are very similar ideas that are fundamentally Sparse Transformer extensions. These models incorporate the notion of a global model memory, which is reminiscent of the Set Transformer's inducing point method or the global model memory of the Star Transformer. Modifications to strides, such as using dilated windows was also proposed in the Longformer work.

The most recent wave of models we've been seeing is models that are based on low-rank approximation or kernel methods, e.g., models such as Low-Rank Transformer~\citep{winata2020lightweight}, Linformer~\citep{wang2020linformer}, Performer~\citep{choromanski2020masked} and/or Linear Transformers~\citep{katharopoulos2020transformers}. Although due to the state of evaluation and the high parallelism of research, it is quite unclear if this low-rank or kernel paradigm is actually better than the learnable pattern (LP) or model memory based efficient Transformer models.  

More recently, there have been more models that propose a two-pronged or two-step attention mechanism combining models from different techniques. The Long Short Transformer~\citep{zhu2021long} is a dynamic form of Linformer combined with Fixed Pattern attention mechanisms. On the other hand, models like Poolingformer also explicitly construct a two-level attention mechanism with techniques reminiscent of memory-based approaches and local attention. Scatter Brain is a new work~\citep{chen2021scatterbrain} attempts to unify sparse (fixed pattern) attention with low-rank attention. Two stage attention mechanisms are also proposed by Luna~\citep{ma2021luna}

On the side, it is important to note that the recurrent based models (Transformer-XL and Compressive Transformers) seem to operate orthogonally and are not as directly comparable to the other models. We also observe that Sparse models~\citep{lepikhin2020gshard,fedus2021switch} that are not only applicable to attention modules, are also recently emerging and becoming more popular and have demonstrated considerable success in the recent months~\citep{du2021glam}.

% Speculate / forecast future trends?

% A healthy line of research based on completely dispensing with attention, i.e., dynamic convolutions~\citep{wu2019pay} or 

\subsection{Brief Discussion on Orthogonal Efficiency Efforts}
While this paper is mainly focused on (1) the computational and memory complexity of the self-attention module and (2) sparsity and adaptive computation, we briefly summarize several orthogonal efforts that may also contribute to model efficiency, scalability, and overall usability of Transformer models. 
\begin{itemize}
\item \textbf{Weight Sharing} - Sharing parameters of the Transformer models would help in reducing overall model size. The Universal Transformers~\citep{dehghani2018universal} tie attention and transition weights across layers. Similarly, Albert~\citep{lan2019albert} does the same parameter sharing across layers. On the other hand, the Quaternion Transformer~\citep{tay2019lightweight} proposes a weight sharing scheme inspired by Hamilton products that locally shares the components in the linear transformation layers.

\item \textbf{Quantization / Mixed Precision} - Learning mixed precision models has the potential to improve memory costs. Q-BERT~\citep{shen2020q} is a model that quantizes Transformer models to ultra-low precision. Meanwhile mixed precision training~\citep{ott2019fairseq} is a highly popular technique to reduced the memory costs of training Transformers.~\citet{fan2020training} applies Quantization Aware training to Transformer models.

\item \textbf{Inference-time Efficiency and Network Pruning} - Multiple research directions explore improving the Transformer efficiency at inference time. One prime example is network model. An example is to prune attention heads during inference~\citep{voita-etal-2019-analyzing,michel2019sixteen}. This has shown to have minimal degradation of performance on downstream tasks. On the other hand, \citet{lagunas2021block} proposes a ``block'' pruning approach which can make a Transformer 2.4x faster with little loss in predictive performance on language tasks. Another line of work involved fast exit during inference which allows us to exit compute if the model is confident of its predictions~\citep{schuster-etal-2021-consistent}. 

\item \textbf{Knowledge Distillation} - 
Knowledge distillation (KD)~\citep{hinton2015distilling} has been a useful technique for transfering the knowledge learned from a larger teacher model to a smaller student model. The smaller model can then be efficiently deployed into production. There have been many attempts to distill large Transformer models. For example, DistilBERT~\citep{sanh2019distilbert}, task-specific distillation~\citep{tang2019distilling} and TinyBERT~\citep{jiao2019tinybert}.

\item \textbf{Neural Architecture Search (NAS)} -
Searching for more efficient Transformer architectures is also a common strategy. \citet{guo2019nat} proposed Neural Architecture Transformer (NAT), using NAS to search for more compact and efficient Transformers by removing redundant operations. \citet{wang2020hat} proposed HAT (Hardware-aware Transformers), a method that leverages NAS and uses hardware efficiency feedback as a reward signal.

\item \textbf{Task Adapters} - This line of research has been primarily focused on the problem of fine-tuning large Transformer on $T$ tasks and aiming to reuse parameters across a variety of tasks. The key idea is that task adapters~\citep{houlsby2019parameter} enable reuse of parameters across tasks and reuse the need of serving $T$ models in production - resulting in overall parameter savings. A modest number of models have been proposed, such as PALS~\citep{stickland2019bert}, MAD-X~\citep{pfeiffer2020mad} and HyperGrid~\citep{tay2020hypergrid}.

\item \textbf{Alternative Architectures} - A considerable amount of effort have gone into designing Transformer alternatives. Amongst the many alternatives considered, a prominent line of emerging research belongs to the family of MLP Mixers~\citep{tolstikhin2021mlp}. Different mixing operations have been proposed, such as the G-MLP~\citep{liu2021pay}, FNet~\citep{lee2021fnet}. Synthesizers~\citep{tay2020synthesizer}, although commonly referred to as an efficient attention method, is also an early manifestation of the mixer line of work, as the random matrices similarly act as an unconditioned mixing operation. A recent promising line of work, based on Structured State Spaces~\citep{gu2021efficiently} also demonstrated very promising results on long range modeling. Lastly, convolutional models are generally more efficient than Transformers since convolutional kernels operate on a fixed, small local neighborhood around the input token. \citet{tay2021pre} shows that, when pre-trained, these more efficient convolutional models can sometimes match the predictive performance of Transformer ones.
\end{itemize}

\subsection{A Retrospective on the Past Year and Future Research Directions}

With our timely V2 update of this survey (updated December 2021), we present retrospective thoughts about how the field has evolved over the past year or so. From the time of last update, it is undeniable that more xformer variants have emerged to offer more efficient alternatives for vanilla Transformers.

Notably, examples of these include  Nystr\"{o}mformer~\citep{xiong2021nystr}, Perceiver~\citep{jaegle2021perceiver}, RFA~\citep{peng2021random}, Luna~\citep{ma2021luna} and Long Short Transformer~\citep{zhu2021long}. There were also other notable models that sprung up around the time when this manuscript was published and narrowly missed the inclusion in the first edition (e.g., Funnel Transformer~\citep{dai2020funnel}). Amongst all the new xformer variants, it is good to note that most do not stray away from the fundamental concepts presented in the first version. Our taxonomy and categorization was more or less broad enough to capture many of these models as they use fundamental ideas that are already present in existing work and therefore can be categorized appropriately. 
Many works can be thought of explicit combinations of existing techniques (two-staged or combination of two method classes) or improvements over existing methods (dynamic formulation of Linformer's low rank projection or better kernels for Linear Transformers). Even though many existing \textit{`memory'} models utilize a form of downsampling to achieve a speed and efficiency gain, we erected a new categoriation of \textit{`downsampling'} to better reflect this new emerging trend~\citep{dai2020funnel,jaegle2021perceiver,tay2021charformer,ryoo2021tokenlearner}.

Over the past year, it is evident that a lot of research investment have been poured into making quadratic attention scalable, in terms of complexity, or sometimes memory.  
At this juncture, it is good to ponder about real tangible need for linear-time attention. Many applications even in language and vision are still dominated by vanilla Transformers with quadratic attention and \emph{none of these xformer variants have caught on as the defacto standard}. There might be multiple explanations from multiple angles for this phenomena. Firstly, linear attention (e.g., Performer) models struggle to be competitive on common benchmarks, as noted from multiple sources~\citep{xiong2021simple,anonymous2021scalinglawmodel}. 

It is good to note that, apart from toy setups or specific domains and problems, they have never been battle-tested against common paradigms like pretrain-and-finetuning only up till recently. Meanwhile, local attention models based on fixed and/or learned patterns such as Sparse Transformers~\citep{child2019generating}, Longformer~\citep{beltagy2020longformer}, ETC~\citep{ainslie2020etc} or BigBird~\citep{zaheer2020big} have seen more reasonable usage, especially within the areas of long context question answering. 
% Swin Transformer~\citep{liu2021swin} has also been widely adopted in the context of computer vision tasks.
However, the high intrinsic implementation complexity of methods such as in ETC~\citep{ainslie2020etc} (substantially increases code complexity by having so many different directions of attention), Swin Transformer~\citep{liu2021swin} or Longformer~\citep{beltagy2020longformer} (requiring custom CUDA kernels and thus making it prohibitive on hardware such as TPUs) might be reasons why these models have yet to found themselves serving as a good, simple-to-use drop-in Transformer replacement.

As noted by~\citep{rabe2021self}, for applications that require to flex on sequence length and memory needs time to time, it might be suffice to \textit{`just sequentially process it'} even if that might not be inherently as satisfying as finding a theoretical approximate. In parallel,~\citep{xiong2021simple} suggests that local attention, when done right, can be a really tough baseline to beat. 

A notable fact about the barrage of efficient attention models is the overloading of the term efficient. It is commonly misunderstood that efficient attention models always imply that the Transformer is fast. The truth is that many of these efficient attention models, owing to their innovation constraints, may make the model much slower. Moreover, many linear attention models do not observe any speed or memory gain at all if the sequence length is short. Many of them have extraordinarily painful requirements to achieve causal masking (or TPU packing)~\citep{choromanski2020rethinking,peng2021random,wang2020linformer} and often have to substantially trade-off throughput for linear complexity. On the other hand, some models cannot be packed or causally masked at all. More notes and discussions about this efficiency misnomer can be found in this paper~\citep{dehghani2021efficiency} which we encourage readers to also peruse.

This update also extends the original scope of efficient attention based xformer models to sparse models even if they did not necessarily target the attention modules. We believe that sparse models were a necessarily addition to the scope to this paper given its recent signs of promise \citep{fedus2021switch,du2021glam,zoph2022designing}. A special note was made to recognize the work done in alternative architectures in the past year (in the section on orthogonal directions). Mixer type architectures \citep{tolstikhin2021mlp} have garnered some interest in computer vision but seem to not perform well on language \citep{anonymous2021remixer}. Meanwhile, alternative models based on Structured State Spaces such as S4 \citep{gu2021efficiently} have solved the hardest Path-X task in the Long Range Arena benchmark \citep{tay2020long}. It should be exciting to see how a model such as S4 would perform at scale, and under pretrained conditions.

As the year comes to a close and as we reflect back on the amazing advances made by the community, we begin to ponder about the future of \textit{efficient transfomers} and what the ideal transformer model should look like. We think that the ideal xformer should hopefully take care of the quadratic memory problem, while retaining universality (e.g., do well on most tasks and not only on long range tasks). The ideal xformer should also not trade-off speed for memory and should not sacrifice the ability to be TPU-packed and/or causal masked. It should ideally be simple and not make use of rigid hard-coding or over-excessive engineering, i.e., it should be elegant and scale well. Ideally, efficiency would be baked right into the next generation of Transformers instead of always having a side variant that one could use for long context tasks. While we cannot explicitly point at any of the xformer variants as the definitive one that have solved the efficiency problem in Transformers, we are optimistic that, given about the pace of advance, \textit{the} true xformer will emerge eventually. It is then a question of whether that new xformer will still be a Transformer. 

\section{Conclusion}
In this paper we surveyed the literature on efficient Transformer models especially pertaining to the quadratic complexity of the self-attention module. We provided a taxonomy and high-level abstraction of the core techniques employed in these class of new models. We characterized the existing models based on techniques and provided a comprehensive walkthrough of several of the efficient Transformer models. Finally, we discussed the evaluation landscape of these models along with the design trends of these models. We ended of with a brief discussion of other parallel orthogonal efforts that may improve the efficiency of Transformer models in general. \textbf{Note:} This survey may be revised again bi-annually or annually. Feel free to send feedback to our email address. While we may not reply to all, we certainly would read them. We also welcome anonymous feedback to \url{https://forms.gle/kqjmhSDEQrmL4Egk6}.

\acks{The authors would like to send the numerous authors who send us feedback via email. We tried our best to incorporate most of the suggestions as we sat fit. We also thank Tamas Sarlos for feedback on this manuscript.}

% Acknowledgements should go at the end, before appendices and references

% \acks{We would like to acknowledge support for this project
% from the National Science Foundation (NSF grant IIS-9988642)
% and the Multidisciplinary Research Program of the Department
% of Defense (MURI N00014-00-1-0637). }


\newpage

\vskip 0.2in
\bibliography{references}

\end{document}